\documentclass[11pt,final]{article}

\usepackage{verbatim}

\usepackage{amsmath}
\usepackage{amssymb}
\usepackage{amsfonts}

\def\@prin{}

\newlength{\atextwidth}
\usepackage[latin1]{inputenc}
\usepackage{esdiff}
\usepackage[dvips]{graphicx}
\usepackage{url}

\usepackage[english]{babel}

\newtheorem{thm}{Theorem}

\newtheorem{defn}{Definition}

\title{\textbf{Total Variation Minimization and Graph Cuts for Moving Objects Segmentation}}
\author{F. Ranchin\footnote{CEREMADE, Université Paris Dauphine, Paris, France; ranchin@ceremade.dauphine.fr}, A. Chambolle\footnote{CMAP, Ecole Polytechnique, Palaiseau, France; antonin.chambolle@polytechnique.fr}, F. Dibos\footnote{LAGA \& L2TI, Universit\'e Paris 13, Paris, France; dibos@math.univ-paris13.fr} 
}

\begin{document}
\maketitle

\begin{abstract}
In this paper, we are interested in the application to video segmentation of the discrete shape optimization problem
\begin{equation} 
\lambda J(\theta)+\sum_{i} (\alpha-f_{i})\theta_{i}
\label{eq:func}
\end{equation}
incorporating a data $f=(f_{i})$ and a total variation function $J$, and where the unknown $\theta=(\theta_{i})$ with $\theta_{i}\in \{0,1\}$ is a binary function representing the region to be segmented and $\alpha$ a parameter. Based on the recent works \cite{TV_MRF}, 
 and Darbon and Sigelle \cite{DS1,DS2}, we justify the equivalence of the shape optimization problem and a weighted TV regularization in the case where $J$ is a ``weighted'' total variation. For solving this problem, we adapt the projection algorithm proposed in \cite{tv_min} to this case. Another way of solving (\ref{eq:func}) investigated here is to use graph cuts. Both methods have the advantage to lead to a global minimum.\\
Since we can distinguish moving objects from static elements of a scene by analyzing norm of the optical flow vectors, we choose $f$ as the optical flow norm. In order to have the contour as close as possible to an edge in the image, we use a classical edge detector function as the weight of the weighted total variation. This model has been used in the former work \cite{ranchin}. We also apply the same methods to a video segmentation model used by Jehan-Besson, Barlaud and Aubert. In this case, it is a direct but interesting application of \cite{TV_MRF}, as only standard perimeter is incorporated in the shape functional. We also propose another way for finding moving objects by using an \emph{a contrario} detection of objects on the image obtained by solving the Rudin-Osher-Fatemi Total Variation regularization problem.We can notice the segmentation can be associated to a level set in the former methods.\\
\end{abstract}

\textbf{Keywords~: total variation, motion detection, active contour models.}

\section{Introduction}
Segmentation of moving objects from a video sequence is an important task whose applications cover domains such like video compression, video surveillance or object recognition. In video compression, the MPEG-4 video coding standard is based on the representation of the scene as different shapes-objects. This representation simplifies the scene and is used for the encoding of the sequence.\\
There are different ways to perform moving objects segmentation, using different mathematical techniques. For Markov Random Fields based methods, we refer to the works of Bouthemy (\cite{BL}, \cite{PHB}) and for maximum likelihood based methods, to the works of Deriche and Paragios (\cite{DerPar}). For variational techniques, we refer to the works of Deriche \emph{et al.} (\cite{ADK}) and Barlaud \emph{et al.} (\cite{AubBarJeh}). At last, mathematical morphology has been more and more used these last ten years, see the works of Salembier, Serra and their teams (\cite{morpho}).\\
In this paper, based on the former work \cite{ranchin} concerning moving object segmentation, we focus on two different techniques, the first one relying on the recent result of \cite{TV_MRF} (the same results were derived independently,
and previously, by Darbon and Sigelle~\cite{DS1,DS2} in a probabilistic setting)
and the second one is the use of graph cuts (Boykov, Veksler, and Zabih \cite{graphcuts}, Kolmogorov and Zabih \cite{gc_klm}).\\ 
The result of \cite{TV_MRF} states that solving the Rudin-Osher-Fatemi Total Variation regularization problem \cite{ROF} and thresholding the result at the level $\alpha$ gives the region that is solution of the shape optimization problem \ref{eq:shapeopti}. The idea of the proof relies on the fact that 
the total variation of a function can be reconstructed from the perimeters
of its level sets:
it is the famous \emph{coarea formula}. Former works rely also on the coarea formula: in \cite{francoise}, the authors propose to use it to propose a new scheme for TV diffusion and improve its efficiency in \cite{francoise2} using a level set decomposition of the image; Chan, Esedoglu and Nikolova in \cite{Nikolova} solve a Mumford-Shah/Chan-Vese (\cite{MS},\cite{CV}) problem with fixed means by a TV-regularization and state also an equivalence result between some special shape optimization problem and a TV regularization one with $L^1$ norm data fidelity term.\\
In this paper,
we use the framework of~\cite{TV_MRF} in the case of a non-homogeneous total
variation functional, corresponding to a weighted anisotropic perimeter like
the one studied in \cite{ranchin}.
 The outline is the following~: in the first part we present the energy used to segment moving objects in the image in the second part and we expose formal mathematic arguments for the use of TV regularization. It is followed by a mathematical part about TV regularization and results about the equivalence with solving a class of shape optimization problems, and by a part where we present graph cuts and their use for our functional. It is followed by an experimental part where we show the results obtained. The last part is dedicated to an automatic moving objects detection performed by \emph{a contrario} statistical methods on the result obtained by total variation regularization (previous parts). We compare it to the previously shown methods.

\section{A shape optimization problem for moving object detection}
\subsection{The functional}
Once we have determined the optical flow, we keep it for the segmentation purpose. We will denote $\Omega$ the moving region and $D$ the image domain. As a moving object should be characterised by a sufficiently large flow magnitude, it seems natural to incorporate $\int_{\Omega} \alpha-|\mathbf{v}|(x)\,dx$ to the energy we want to minimize, where $\alpha-|\mathbf{v}|(x)$ have to take different signs on the image domain, otherwise the solution of the shape optimization problem will be trivial. As we want the boundary of $\Omega$ 
to remain stable in the presence of noise or spurious variations, we also
penalize the total length of this boundary (that is, the perimeter of $\Omega$)
in our functional.
Finally, as thresholding the optical flow will not give exact object contours (due to the temporal integration), we add a weighted perimeter which integrates a function of the gradient (here $g_{I}=\frac{1}{1+|\nabla I|^2}$) along the boundary. It gives the functional 
\begin{equation}
\int_{\Omega} \alpha\,dx+\int_{D\setminus\Omega} |\mathbf{v}|\,dx+\lambda\int_{\partial \Omega} g_{I}(x)\,dS+\mu\int_{\partial \Omega} \,dS
\label{eq:functional}
\end{equation}
where $dS$ denotes the arclength variation along the boundary. For simplicity notations, we will denote $\lambda g_{I}+\mu$ by $g$. Finally, our functional is
\begin{equation}
\int_{\Omega} \alpha\,dx+\int_{D\setminus\Omega} |\mathbf{v}|\,dx+\int_{\partial \Omega} g(x)\,dS
\label{eq:energy}
\end{equation}
Within the framework of shape sensitivity analysis (see
Murat and Simon~\cite{MuratSimon}, Delfour and Zolesio \cite{DelZol}), 
one can compute the \emph{shape derivative} of this functional and obtain the steepest gradient descent. Combining it to the famous level set method (Osher, Sethian, \cite{OS}), we would obtain
$$\diffp{u}{t}=|\nabla u|\left(|\mathbf{v}|-\alpha+\mathrm{div}\left(g\frac{\nabla u}{|\nabla u|}\right)\right).$$
Another similar method is to use $u$ as the unknown of the functional and not $\Omega$~: the integral over $\Omega$ (\emph{resp.} $D\setminus\Omega$) is replaced by integrals over $D$ with the weight $H_{\epsilon}(u)$ (\emph{resp.} $1-H_{\epsilon}(u)$) and the boundary term by the integral over $D$ with the weight $|\nabla (H_{\epsilon}(u))|$. Let notice that a parameter $\epsilon$ is needed in this method for computing $\delta_{\epsilon}$ and $H_{\epsilon}$ which are $C^{\infty}$ regularizations of Dirac and Heaviside distributions. The obtained PDE, leading to the same curve motion than the previous one, is
$$\diffp{u}{t}=\delta_{\epsilon}(u)\left(|\mathbf{v}|-\alpha+\mathrm{div}\left(g\frac{\nabla u}{|\nabla u|}\right)\right).$$
That was done in \cite{ranchin}, unfortunately, if we want to adjust the value of $\alpha$ in a suitable way, we have to recompute the result by this partial differential equation as many times as necessary. We overcome this problem by using the equivalence between solving the ROF model with a weighted total variation and solving of (\ref{eq:functional}) for all the possible values of $\alpha$.\\
In \cite{TV_MRF}, functionals do not involve standard perimeter but a different anisotropic one. This is for theoretical reasons explained in \cite{TV_MRF}~: the discrete total variation does not satisfy the coarea formula which is needed in the main result of \cite{TV_MRF}.
In fact, the theory can be developped with the isotropic total variation
in the continuous setting, and results could still be (approximately) computed.
\\
Thus we slightly modify the functional to fit in the framework given in \cite{TV_MRF} ($\nu$ denotes the outside normal to the boundary and $|\cdot|_{1}$ the 1-norm~: $|(a,b)|_{1}=|a|+|b|$, $R_{\frac{\pi}{4}}$ denotes the rotation of angle $\frac{\pi}{4}$)
\begin{equation}
E(\Omega)=\int_{\Omega} \alpha\,dx+\int_{D\setminus\Omega} |\mathbf{v}|\,dx+\frac{1}{2}\int_{\partial \Omega} g(x)(|\nu|_{1}+|R_{\frac{\pi}{4}}(\nu)|_{1})\,dS.
\label{eq:my_energy}
\end{equation}
This is a change of metric~: the standard length and its weighted counterpart are replaced by what it is usually called ``Manhattan'' or ``taxicab'' length.
We could keep only $\int_{\partial \Omega} g(x)|\nu|_{1}\,dS$ but $\int_{\partial \Omega} \frac{1}{2}g(x)(|\nu|_{1}+|R_{\frac{\pi}{4}}(\nu)|_{1})\,dS$ is useful to not overestimate the length of diagonal linear parts of the boundary of $\Omega$.
We introduce the weighted isotropic and anisotropic total variations
\[
TV_{g}(u)\ :=\ \int_D g|Du| \ \textrm{ and }
TV_{1,g}(u)\ :=\ \frac{1}{2}\int_D g(|Du|_1+|R_{\frac{\pi}{4}}(Du)|_1)\,,
\]
(notation 1 refers to the $1$-norm of the normal and $g$ to the weight function)
so that $TV_{1,g}(\chi_{\Omega})=\int_{\partial \Omega} \frac{1}{2}g(x)(|\nu|_{1}+|R_{\frac{\pi}{4}}(\nu)|_{1})\,dS$ and
$TV_{g}(\chi_{\Omega})=\int_{\partial \Omega} g(x)\,dS$ are
respectively the anisotropic weighted perimeter and the weighted perimeter. 
We denote
$\Lambda_{g}(\partial\Omega)=TV_{1,g}(\chi_{\Omega})$ and $L_{g}(\partial\Omega)=TV_{g}(\chi_{\Omega})$: these two perimeters satisfy
$$c_{1}L_{g}(\partial\Omega)\leq\Lambda_{g}(\partial\Omega)\leq c_{2}L_{g}(\partial\Omega)$$
with $c_{1}=\frac{1+\sqrt{2}}{2}$, $c_{2}=\frac{1}{\sqrt{2-\sqrt{2}}}$, and thus if the boundary of $\Omega$ has a finite $L_{g}$, it has finite $\Lambda_{g}$, and conversely.\\
At last, we rewrite our functional in discrete setting, as this will be in the rest of the paper
\begin{multline*}
E(\theta)=\sum_{i,j} (\alpha -|\mathbf{v}|_{i,j})\theta_{i,j}+\frac{1}{2}\sum_{i,j} g_{i,j}\left(|\theta_{i+1,j}-\theta_{i,j}|+|\theta_{i,j+1}-\theta_{i,j}|\right)\\
+\frac{1}{2\sqrt{2}}\sum_{i,j} g_{i,j}\left(|\theta_{i+1,j+1}-\theta_{i,j}|+|\theta_{i-1,j+1}-\theta_{i,j}|\right).
\end{multline*}
Let us observe that the weight $g_{i,j}$ could be different on each
edge (connecting two neighboring pixels) of the grid and that the choice
we have made is quite arbitrary. However, we did not observe a significant
change in the output when weighing the edges in a different way.
\subsection{Remarks about the minimization}
As we have seen, a functional like (\ref{eq:energy}) is usually minimised using shape sensitivity analysis \cite{MuratSimon,zolesio,DelZol}, classical calculus of variation (see for example \cite{CKS}) 
or heaviside function techniques (Chan-Vese, \cite{CV}). All of those are gradient-descent methods. In \cite{TV_MRF,DS1,DS2}, it is shown that the solutions of the discrete shape optimization problem
$$\min_{\theta,\,\theta_{i}\in\{0,1\}} \lambda J(\theta)+\sum_{i} (\alpha-f_{i})\theta_{i}$$
($i$ is an index of the pixel number and $\theta$ plays the role of the characteristic function of the shape, $f$ is a data function [in our problem it is the optical flow norm] and $J$ is a total variation, though it could be another function satisfying the same properties, this will be described in section \ref{sec:TVtheory}) can be obtained by computing the solution of the Rudin-Osher-Fatemi total variation regularization problem
$$\min_{u} \frac{1}{2\lambda}\Vert u-f\Vert^2+J(u)$$
and just threshold the result $\tilde{u}$ at the level $\alpha$. This has two advantages over classical snakes methods like the ones cited above. First, it gives a global minimum of the shape optimization problem, which is not necessarily the case of the classical snakes methods, since the gradient descent may be trapped into local minima. Secondly, if we want to find the most appropriate value of $\alpha$, we have just to compute \textit{once} the solution of the ROF problem and to threshold at different levels in order to decide the value we keep; by any other method, we would be obliged to repeat the minimization as many times as the number of values of $\alpha$ we would like to compare. With the projection algorithm for computing the solution of the ROF problem (see section \ref{sec:proj}), we inherit of another slighter advantage~: we avoid introducing additional parameters which are required to approximate either the total variation in usual solving by PDE, or Dirac and Heaviside functions (see \cite{CV} for details).\\
It is known since Greig, Portehous ans Seheult \cite{GPS} that energies (\ref{eq:energy}) and (\ref{eq:my_energy}) can be exactly minimized. More recently, Kolmogorov and Zabih in \cite{gc_klm} proposed the "`graph cuts"' algorithm as a way to minimize such type of energies. We will detail about it in the section \ref{sec:graphcuts}. It leads to a global minimum, but the second advantage of TV regularization does not occur here~: we have to compute the solution of the shape optimization problem as many times as necessary if we want to optimize the $\alpha$ parameter. As a single graph cut computation requires approximately $0.5$ second and the ROF solution about $1$ minute (on an image of size $256\times 256$ on a laptop equiped with a 1.8 GHz Pentium 4 and 1 Gb of RAM), graph cuts are better for a computation for a fixed value of $\alpha$, but if we want to choose many different values of $\alpha$, the ROF solution computation should be more indicated.

\section{On the equivalence of total variation regularization and a class of shape optimization problems}
\label{sec:TVtheory}
In this section, we will use the following notations~: $|\cdot|$ denotes the euclidean norm $|(a,b)|=\sqrt{a^2+b^2}$, $|\cdot|_{p}$ denotes the $p$-norm $|(a,b)|_{p}=(|a|^{p}+|b|^{p})^{1/p}$ and $|\cdot|_{\infty}$ denotes the $\infty$-norm $|(a,b)|_{\infty}=\sup(|a|,|b|)$
\subsection{Settings}
In this section, we recall the main results obtained in \cite{TV_MRF}. The problem considered is
$$\min_{\theta\in X,\theta_{i}\in\{0,1\}} \lambda J(\theta)+\sum_{i} (\alpha-f_{i})\theta_{i} \qquad (P_{\alpha})$$
where $X$ is the space of functions defined on the $N$ pixels of the image grid ($i$ denotes the pixel index and $f$ is still a data function).
The function $J:X\to \mathbb{R}^{+}$ satisfies four properties. 
\begin{itemize}
\item Convexity~: $J(tu+(1-t)v)\leq tJ(u)+(1-t)J(v)$ for any $u,v \in X$ and $t\in[0,1]$,
\item lower semicontinuity,
\item $1$-homogeneity~: $J(tu)=tJ(u)$ for any $t\geq 0$ and $u\in X$,
\item it satisfies also the generalized \emph{co-area formula}
\begin{equation}
J(u)=\int_{-\infty}^{+\infty} J(\mathbf{1}_{u>t})\,dt
\label{eq:coarea}
\end{equation}
where $\mathbf{1}_{u>t}$ denotes the indicator function of the upper level set of $u$.
\end{itemize}
\subsubsection{Main theorem and extensions}
We consider the Rudin-Osher-Fatemi TV regularization problem
\begin{equation}
\min_{u\in X} J(u)+\frac{1}{2\lambda}\Vert u-f\Vert^2
\label{eq:ROF}
\end{equation}
and the discrete shape optimization problem
\begin{equation}
\min_{\theta\in X, \theta_{i}\in\{0,1\}} \lambda J(\theta)+\sum_{i} (\alpha-f_{i})\theta_{i}
\label{eq:shapeopti}
\end{equation}
The main theorem of \cite{TV_MRF} states an equivalence between solving (\ref{eq:ROF}) and thresholding the result at threshold $\alpha$ and solving (\ref{eq:shapeopti}). As we are concerned only with solving (\ref{eq:shapeopti}), we give only the part of the theorem which states that thresholding the solution of the discretized ROF model gives a solution of the shape optimisation problem.
\begin{thm} {\bf (\cite{TV_MRF})}
Let $w$ solve (\ref{eq:ROF}). Then, for any $s\in\mathbb{R}$, both $w_{i}^{s}=\mathbf{1}_{w_{i>s}}$ and $\bar{\bar{w}}_{i}^{s}=\mathbf{1}_{w_{i>s}}$ solve (\ref{eq:shapeopti}). If $w^{s}=\bar{\bar{w}}^{s}$, then the solution of (\ref{eq:shapeopti}) is unique.
\label{th:chambolle}
\end{thm}
In \cite{TV_MRF}, it is the discrete Manhattan total variation that is used 
$$J(u)=\sum_{i,j} |u_{i+1,j}-u_{i,j}|+|u_{i,j+1}-u_{i,j}|$$
which is dicretized from the continuous $1$-TV introduced in the previous section. If we want a more isotropic and $\frac{\pi}{4}$-rotationnally invariant Manhattan TV, we may take diagonal terms into account
$$\frac{1}{2}\sum_{i,j} |u_{i+1,j}-u_{i,j}|+|u_{i,j+1}-u_{i,j}|+\frac{1}{2\sqrt{2}}\sum_{i,j} |u_{i+1,j+1}-u_{i,j}|+|u_{i-1,j+1}-u_{i,j}|,$$
which is discretized from $\frac{1}{2}\int_{D} |\nabla u|_{1}+\frac{1}{2}\int_{D} |\nabla u\cdot e_{1}|+|\nabla u\cdot e_{2}|$ where $e_{1}=(\frac{\sqrt{2}}{2},\frac{\sqrt{2}}{2})$ and $e_{2}=e_{1}^\perp$. The second term can be seen as a Manhattan TV in another basis, actually it is exactly $\int_{D} |R_{\frac{\pi}{4}}(\nabla u)|_{1}$ where $R_{\frac{\pi}{4}}$ is the rotation of angle $\frac{\pi}{4}$.
The discrete standard TV 
$$TV_{1,g}(u)=\sum_{i,j} \sqrt{|u_{i+1,j}-u_{i,j}|^2+|u_{i,j+1}-u_{i,j}|^2}$$
do not fit in the frame described here since it does not satisfy the generalized coarea formula, though being the discretized version of the total variation in the standard definition given in the previous section. \\
As the Theorem \ref{th:chambolle} is stated for any function $J$ satisfying the four conditions given above and the Manhattan discrete TV satisfy them. It is straightforward to extend it to a g-weighted Manhattan TV 
$$\sum_{i,j} g_{i,j}\left(|u_{i+1,j}-u_{i,j}|+|u_{i,j+1}-u_{i,j}|\right)$$
then to the more isotropic
\begin{multline}
TV_{1,\frac{\pi}{4},g}(u)=\frac{1}{2}\sum_{i,j} g_{i,j}\left(|u_{i+1,j}-u_{i,j}|+|u_{i,j+1}-u_{i,j}|\right)\\
+\frac{1}{2\sqrt{2}}\sum_{i,j} g_{i,j}\left(|u_{i+1,j+1}-u_{i,j}|+|u_{i-1,j+1}-u_{i,j}|\right)
\label{eq:tv1gp}
\end{multline}
in which we are concerned in this paper.

\subsection{The projection algorithm of \cite{tv_min}}
\label{sec:proj}
In \cite{tv_min}, a new algorithm for computing the solution of (\ref{eq:ROF}) was proposed. It is based on duality results and consists in finding the projection of $f$ onto a convex set. Let us describe how it works on the energy we are interested in. Here we follow the calculus of \cite{TV_MRF} which generalize well to the g-weighted Manhattan TV\\
The energy considered is thus
$$TV_{1,\frac{\pi}{4},g}(u)=\frac{1}{2}\sum_{i,j} g_{i,j}\left(|(\nabla^{x} u)_{i,j}|+|(\nabla^{y} u)_{i,j}|\right)+\frac{1}{2}\sum_{i,j} g_{i,j}\left(|(\nabla^{xy} u)_{i,j}|+|(\nabla^{yx} u)_{i,j}|\right)$$
where we have rewritten the expression of \ref{eq:tv1gp}. By now, we denote $\nabla w=(\nabla^{x} w,\nabla^{y} w)$ and $\nabla' w=(\nabla^{xy} w,\nabla^{yx} w)$.\\
From discrete gradients, we get the definition of discrete divergence  $\mathrm{div}=-\nabla^{*}$
$$(\mathrm{div} \xi,w)_{X}=-(\xi,\nabla w)_{X\times X},\ \forall w\in X,\xi\in X\times X,$$
and similarly with is rotated counterpart $\mathrm{div}'=-(\nabla')^{*}$
$$(\mathrm{div}' \xi,w)_{X}=-(\xi,\nabla' w)_{X\times X},\ \forall w\in X,\xi\in X\times X,$$

In \cite{TV_MRF}, it is stated that the solution of
$$\min_{w\in X} \sum_{i,j} |(\nabla w)_{i,j}|+\frac{1}{2\lambda} \Vert w-w_{0}\Vert^2$$
(where $|(\nabla w)_{i,j}|$ is the euclidean norm of $(\nabla w)_{i,j}$) is given by $\bar{w}=w_{0}-\lambda\mathrm{div} \bar{\xi}$ where $\bar{\xi}$ is a solution to
$$\min\{\Vert \lambda\mathrm{div} \xi-w_{0}\Vert^2|\,\xi\in X\times X,\ |\xi|\leq 1\}.$$
Moreover, one has $\bar{\xi}_{i,j}\cdot(\nabla \bar{w})_{i,j}=|\nabla \bar{w}|_{i,j}$ for all $(i,j)$. As this duality problem relies on the property
$$\xi\cdot\nabla w\leq |\xi|_{p}|\nabla w|_{q}$$
with $\frac{1}{p}+\frac{1}{q}=1$ with $p\in[1,+\infty]$ (for $p=\infty$, $q=1$ and conversely) and as we have $q=1$ for Manhattan TV, the constraint $|\xi_{i,j}|\leq 1$ is replaced by $|\xi_{i,j}|_{\infty}\leq 1$, that is to say $|\xi^{x}_{i,j}|\leq 1$ and $|\xi^{y}_{i,j}|\leq 1$. For $g$-``weighted'' Manhattan TV, as we want to realize
$$\xi\cdot\nabla w\leq |\xi|_{\infty}|\nabla w|_{1}\leq g|\nabla w|_{1},$$
the constraints become  $|\xi^{x}_{i,j}|\leq g_{i,j}$ and $|\xi^{y}_{i,j}|\leq g_{i,j}$. If we consider the full $TV_{1,\frac{\pi}{4},g}$, we have the part of Manhattan TV expressed in the basis $(e_{1},e_{2})$. This leads to another vector field $\eta$ wich satisfies the same properties as $\xi$. All the constraints can be renormalized by the function $g$ equivalently~: we replace $\mathrm{div}(\xi)$ by $\mathrm{div}(g\xi)$ and $|\xi|\leq g$ by $|\xi|\leq 1$, and for $\eta$ in the same way.
Let introduce the compact set (the overlining denotes the closure)
$$K=\overline{\{\mathrm{div}(g\,\xi)+\mathrm{div}'(g\,\eta)|\,(\xi,\eta)\in A^2\}}$$
where
$$A=\{p=(p^{x},p^{y})\in X\times X,\, |p^{x}_{i,j}|\leq 1,\ |p^{y}_{i,j}|\leq 1\}.$$
From the definition of the total variation $TV_{1,\frac{\pi}{4},g}$, we get 
$$TV_{1,\frac{\pi}{4},g}(w)=\sup_{|\xi|_{\infty}\leq 1} \left(w,\mathrm{div}(g\,\xi)\right)_{X}+\sup_{|\eta|_{\infty}\leq 1} \left(w,\mathrm{div}'(g\,\eta)\right)_{X}=\sup_{v\in K} \left(w,v\right)_{X}.$$
Exactly in the same manner than in \cite{TV_MRF}, it can be established that the solution of the ROF problem is given by the orthogonal projection of $w$ onto $\lambda K$
This is for constraints simplicity. Finally, the solution of 
\begin{equation}
\min_{w\in X} TV_{1,\frac{\pi}{4},g}(w)+\frac{1}{2\lambda} \Vert w-w_{0}\Vert^2
\label{eq:ROF_myTV}
\end{equation}
is given by $\bar{w}=w_{0}-\frac{1}{2}\left(\lambda\mathrm{div}(g\,\bar{\xi})+\lambda\mathrm{div}'(g\,\bar{\eta})\right)$ where $(\bar{\xi},\bar{\eta})$ is a solution to
\begin{equation}
\min_{(\xi,\eta)\in A^2} \Vert \frac{1}{2}\left(\lambda\mathrm{div}(g\,\xi)+\lambda\mathrm{div}'\, (g\eta)\right)-w_{0}\Vert^2
\label{eq:ROFproj}
\end{equation}
Let us mention that the $\mathrm{div}'$ operator is different from the $\mathrm{div}$ one as it is the conjugate of the gradient in the basis $(e_{1},e_{2})$. It is simply given by (denoting $f=(f^{1},f^{2})$) $$(\mathrm{div}'f)_{i,j}=\frac{1}{\sqrt{2}}(f^{1}_{i,j}-f^{1}_{i-1,j+1}+f^{2}_{i,j}-f^{2}_{i-1,j+1}).$$
The Karush-Kuhn-Tucker conditions yield the existence of Lagrange multipliers $\alpha^{1}_{i,j}\geq 0$, $\alpha^{2}_{i,j}\geq 0$, $\beta^{1}_{i,j}\geq 0$, $\beta^{2}_{i,j}\geq 0$ associated tot the constraints in (\ref{eq:ROFproj}) that are $(\xi^{1}_{i,j})^2\leq 1$, $(\xi^{2}_{i,j})^2\leq 1$, $(\eta^{1}_{i,j})^2\leq 1$, $(\eta^{2}_{i,j})^2\leq 1$. These Lagrange multipliers satisfy

\begin{align*}
-\lambda g_{i,j}\nabla(\frac{\lambda}{2}(\mathrm{div}(g\xi)+\mathrm{div}'(g\eta))-w_{0})_{i,j}+2(\alpha^{1}_{i,j}\xi^{1}_{i,j},\alpha^{2}_{i,j}\xi^{2}_{i,j})^{T} & =0\\
-\lambda g_{i,j}\nabla'(\frac{\lambda}{2}(\mathrm{div}(g\xi)+\mathrm{div}'(g\eta))-w_{0})_{i,j}+2(\beta^{1}_{i,j}\eta^{1}_{i,j},\beta^{2}_{i,j}\eta^{2}_{i,j})^{T} & =0 
\end{align*}
with either $\alpha^{1}>0$ (and similarly for $\alpha^{2}$, $\beta^{1}$ and $\beta^{2}$) and $\xi^{1}$. Thus
$$
\begin{array}{ccc}
\alpha^{1}_{i,j} & = & \frac{1}{2}\lambda g_{i,j}|\nabla^{x}(\frac{\lambda}{2}(\mathrm{div}(g\xi)+\mathrm{div}'(g\eta))-w_{0})|\\
\alpha^{2}_{i,j} & = & \frac{1}{2}\lambda g_{i,j}|\nabla^{y}(\frac{\lambda}{2}(\mathrm{div}(g\xi)+\mathrm{div}'(g\eta))-w_{0})|\\
\beta^{1}_{i,j} & = & \frac{1}{2}\lambda g_{i,j}|\nabla^{xy}(\frac{\lambda}{2}(\mathrm{div}(g\xi)+\mathrm{div}'(g\eta))-w_{0})|\\
\beta^{2}_{i,j} & = & \frac{1}{2}\lambda g_{i,j}|\nabla^{yx}(\frac{\lambda}{2}(\mathrm{div}(g\xi)+\mathrm{div}'(g\eta))-w_{0})|\\
\end{array}
$$
Then, we obtain a fixed-point algorithm similar to the one proposed in \cite{TV_MRF}
$$
\begin{array}{lcc}
w^{n} & = & \frac{1}{2}\left(\lambda\mathrm{div}(g\,\xi_{n})+\lambda\mathrm{div}'(g\,\eta_{n})\right)-w_{0}\\
(\xi^{n+1}_{i,j})^{x} & = & \frac{(\xi^{n}_{i,j})^{x}+g_{i,j}\frac{\tau}{\lambda}(\nabla^{x} w^{n})_{i,j}}{1+g_{i,j}\frac{\tau}{\lambda}|(\nabla^{x} w^{n})_{i,j}|}\\
\vspace{0.15cm}
(\xi^{n+1}_{i,j})^{y} & = & \frac{(\xi^{n}_{i,j})^{y}+g_{i,j}\frac{\tau}{\lambda}(\nabla^{y} w^{n})_{i,j}}{1+g_{i,j}\frac{\tau}{\lambda}|(\nabla^{y} w^{n})_{i,j}|}\\
\vspace{0.15cm}
(\eta^{n+1}_{i,j})^{x} & = & \frac{(\eta^{n}_{i,j})^{x}+g_{i,j}\frac{\tau}{\lambda}(\nabla^{xy} w^{n})_{i,j}}{1+g_{i,j}\frac{\tau}{\lambda}|(\nabla^{xy} w^{n})_{i,j}|}\\
\vspace{0.15cm}
(\eta^{n+1}_{i,j})^{y} & = & \frac{(\eta^{n}_{i,j})^{y}+g_{i,j}\frac{\tau}{\lambda}(\nabla^{yx} w^{n})_{i,j}}{1+g_{i,j}\frac{\tau}{\lambda}|(\nabla^{yx} w^{n})_{i,j}|}\\
\end{array}
$$
Following the convergence proof of \cite{tv_min}, we obtain the convergence theorem
\begin{thm}
Let $\tau\leq \frac{1}{8\max_{i,j} g_{i,j}}$. Then, $\lambda\mathrm{div}(g\xi^{n})+\lambda\mathrm{div}'(g\eta^{n})$ converges to the orthogonal projection of $w_{0}$ onto the convex set $\lambda K$ as $n\to\infty$, and $w^{n}$ converges to the solution of (\ref{eq:ROF_myTV}).
\end{thm}

\section{How to minimize the energies with graphcuts}
\label{sec:graphcuts}
\subsection{Principle}
Greig, Portehous and Seheult proved  in \cite{GPS} that discrete energy minimization can be exactly performed. Graphcuts have been introduced in computer vision by Y. Boykov and his collaborators in \cite{graphcuts} as an algorithm for this type of minimization. They have been extended to many areas~: stereovision \cite{gc_stereo}, medical imaging \cite{gc_med}... The idea is to add a ``source'' and a ``sink'' in such a way that to each point in the image grid a link is created to either the source or the sink. A cost is assigned to the links so that the global cost be related to the energy. Finally, solving the energy minimization problem is equivalent to find a cut of minimal cost along the graph (source-points-sink). This is achieved by finding a ``maximal flow'' along the edges of the graph, due to a duality between min-cut and max-flow problems, first observed by Ford and Fulkerson.
\subsection{Construction}
We recall the energy is (we replace $\lambda+\mu g$ by $g$ for simplicity)
$$
\begin{array}{ccr}
J(\theta) & = &\sum_{(i,j)} (\alpha-|\mathbf{v}|_{i,j})\theta_{i,j}+\frac{1}{2}\sum_{i,j} g_{i,j}\bigg(|\theta_{i+1,j}-\theta_{i,j}|+|\theta_{i,j+1}-\theta_{i,j}|\\
& & +\frac{\sqrt{2}}{2}|\theta_{i+1,j+1}-\theta_{i,j}|+\frac{\sqrt{2}}{2}|\theta_{i+1,j-1}-\theta_{i,j}|\bigg)
\end{array}
$$
which gives, with simpler notations (we denote a pixel $x=(i,j)$)
$$
J(\theta)=\sum_{x} (\alpha-|\mathbf{v}|_{x})\theta_{x}+\sum_{x,y} w_{x,y}|\theta_{y}-\theta_{x}|
$$
The coefficients $w_{x,y}$ are given by $w((i,j),(i\pm 1,j))=w((i,j),(i,j\pm 1))=g_{(i,j)}$ and $w((i,j),(i\pm 1,j\pm 1))=w((i,j),(i\mp 1,j\pm 1))=\frac{\sqrt{2}}{2} g_{(i,j)}$.\\
One can see that the weights $w_{x,y}$ are nonsymmetric~: $w_{x,y}\neq w_{y,x}$ due to the presence of $g$ which has a dependency with respect to the pixel. However there is no particular problem to introduce nonsymmetric weights, as Kolmogorov and Zabih have shown in \cite{gc_klm} that graphcuts can handle energies involving an interaction term which satisfies $E_{inter}(0,0)+E_{inter}(1,1)\leq E_{inter}(0,1)+E_{inter}(1,0)$.\\
Then, we build the graph $\mathcal{G}=(\mathcal{V},\mathcal{E})$ made of vertices
$\mathcal{V}=\{i,\,i=1,...,N\}\cup\{t\}\cup\{s\}$
and whose edges are
$$\mathcal{E}=\{(x,y)|\,w_{x,y}>0\}\cup\{(s,x)|\,1\leq x\leq N\}\cup\{(x,t)|\,1\leq x\leq N\}.$$
As a cut of this graph define a partition $(\mathcal{V}_{s},\mathcal{V}_{t})$ of the graph into two sets, the first one containing the source and the second one containing the sink, the global cost of a cut is given by
$$E(\mathcal{V}_{s},\mathcal{V}_{t})=\sum_{\stackrel{e=(a,b)\in\mathcal{E}}{a\in\mathcal{V}_{s},b\in\mathcal{V}_{t}}} C(e).$$
So what we would like to realize is $E(\mathcal{V}_{s},\mathcal{V}_{t})=J(\theta)$. The construction is given by Kolmogorov in \cite{gc_klm}, it consists in assigning  the weight $w_{x,y}$ to an edge $e=(x,y)\in\mathcal{E}$ in the image grid, the weight $\alpha+\max_{i} G_{i}$ to the edges $(s,x)$ and $\max_{i} G_{i} -G_{i}$ to the edges $(x,t)$, then the equality between the global cost and the energy holds.

\section{Experimental results}
All the experiments whose results are presented here were performed on a laptop equiped with a 1.8GHz Pentium 4 and 1 Gb of RAM.
\subsection{Experiments with optical flow}
For the implementation, we have used the \texttt{maxflow-v2.1} and \texttt{energy-v2.1} graphcuts implementation of V. Kolmogorov, available at \url{http://www.cs.cornell.edu/People/vnk/software.html}. Type of capacities has been set to \texttt{double}, though \texttt{short} or \texttt{int} leads to faster computation when quantized quantities are chosen in input.\\ 
The optical flow is computed by the Weickert and Schnörr method \cite{WS} with a multiresolution procedure (see \cite{MP}). As optical flow computation has been improved since the Weickert and Schnörr spatiotemporal model (using mixed model combining local and global information, using intensity or gradient intensity...), we emphasize that our purpose is not to obtain a very precise estimation of the optical flow but to show how we can improve this with the $g$-weighted term and thus to obtain a segmentation as close as possible to the image edges. 
Figure 1 shows results obtained successively with $TV_{1,g}$, $TV_{1,g,\frac{\pi}{4}}$ and weighted standard perimeter $TV_{g}$. One can see the result obtained with Manhattan perimeter (diagonal neighbors) is quite competitive with the one obtained with standard perimeter, especially it is more isotropic, which is precisely what is aimed. Parameters are chosen from previous computations with classical snakes (see \cite{ranchin}). The values are set in relation with the range of value of the optical flow amplitude. 
\begin{figure}
\begin{center}
\includegraphics[width=0.4\textwidth]{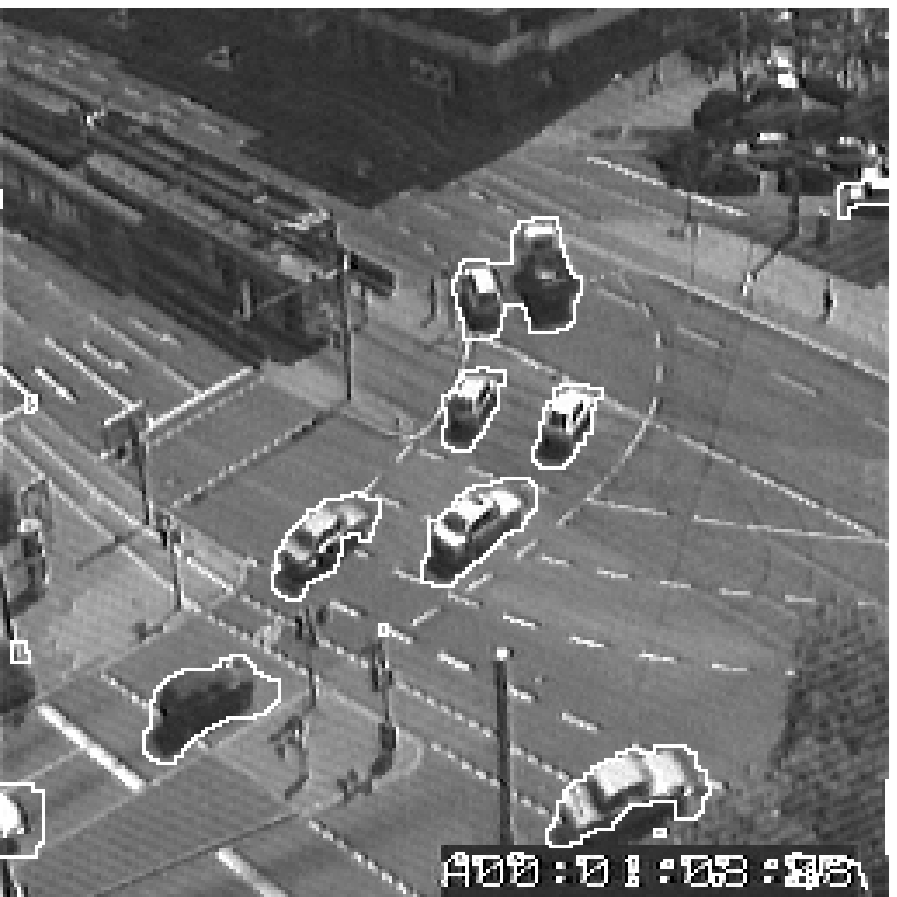}
\includegraphics[width=0.4\textwidth]{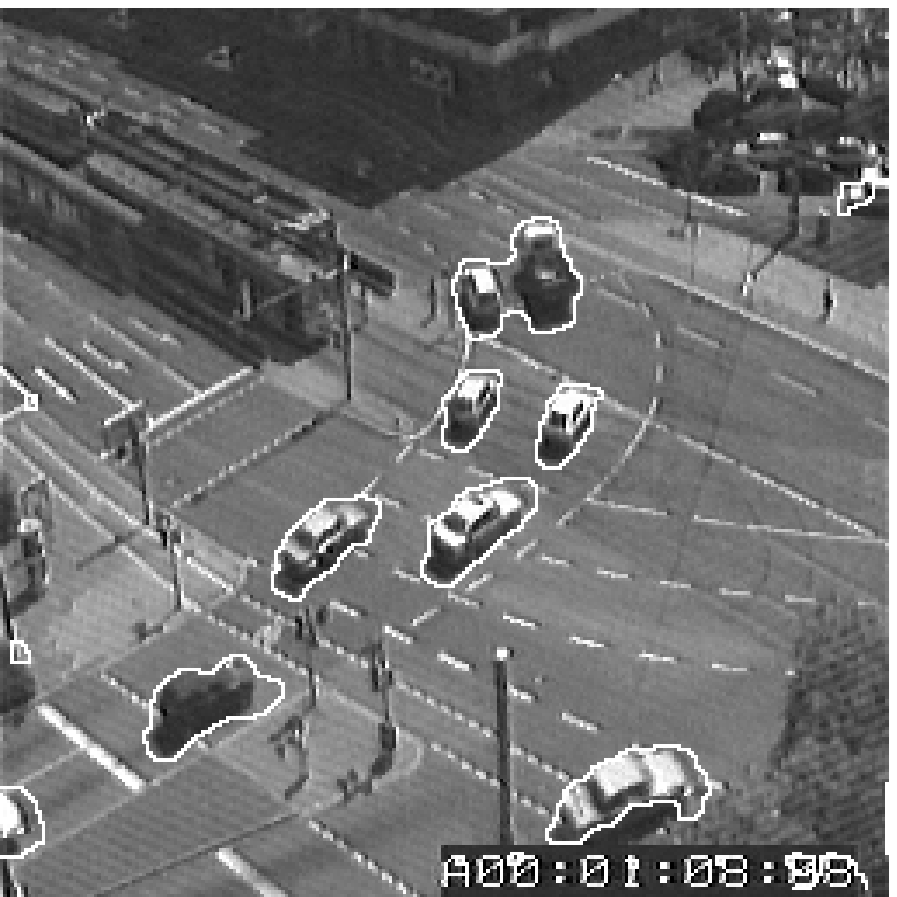}
\includegraphics[width=0.4\textwidth]{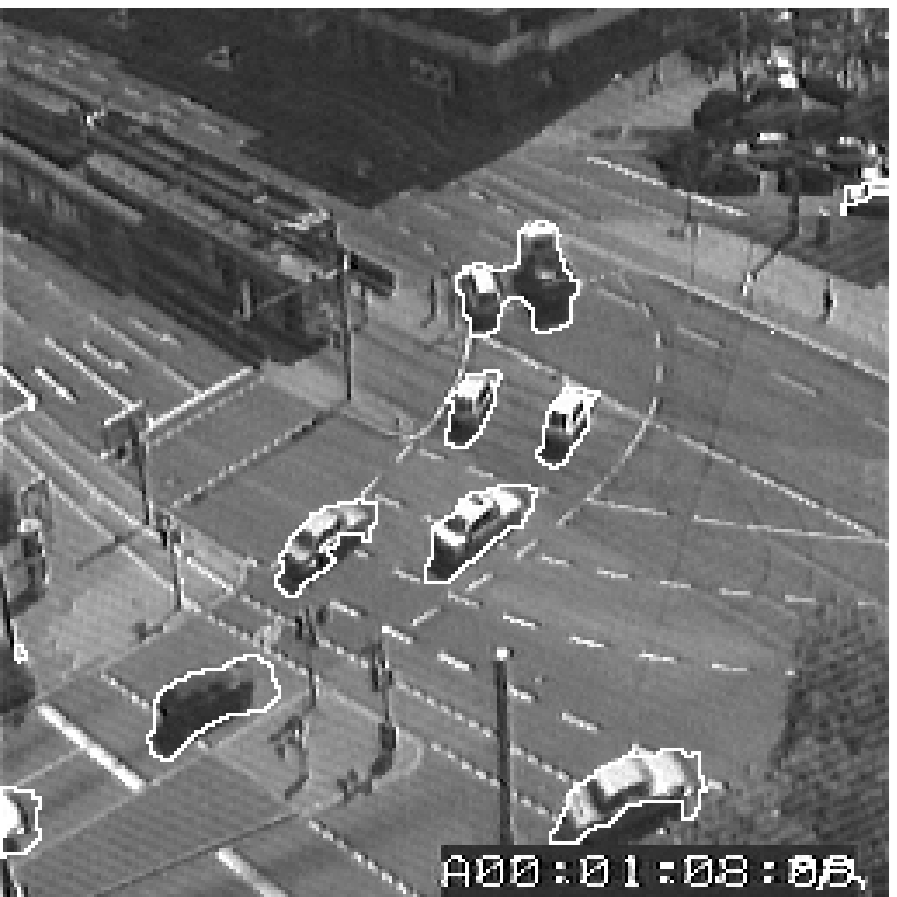}
\end{center}
\caption{{\small Results obtained with graphcuts with the energy
    involving $TV_{1,g}$ (first image on top left), $TV_{1,g,\frac{\pi}{4}}$ (top right) and $TV_{g}$ (bottom). The initial data is the optical flow norm $\mathbf{v}$. Parameters are $\alpha=0.6$, $\lambda=0.2$ and $\mu=10$.}}
\label{fig:toto}
\end{figure} 
For the weighted standard perimeter, the result is obtained in $0.24$ or $0.25$ second on all the images (of size $256\times 256$) of the sequence. For weighted Manhattan perimeter involving diagonal neighbors, the time is of $0.11$, $0.12$ or $0.13$ second. Same times are obtained with weighted Manhattan perimeter, though it can reach $0.09$ or $0.10$ second on some images. All of these are obtained with the \texttt{clock()} C command. 
\begin{figure}
\begin{center}
\includegraphics[width=0.4\textwidth]{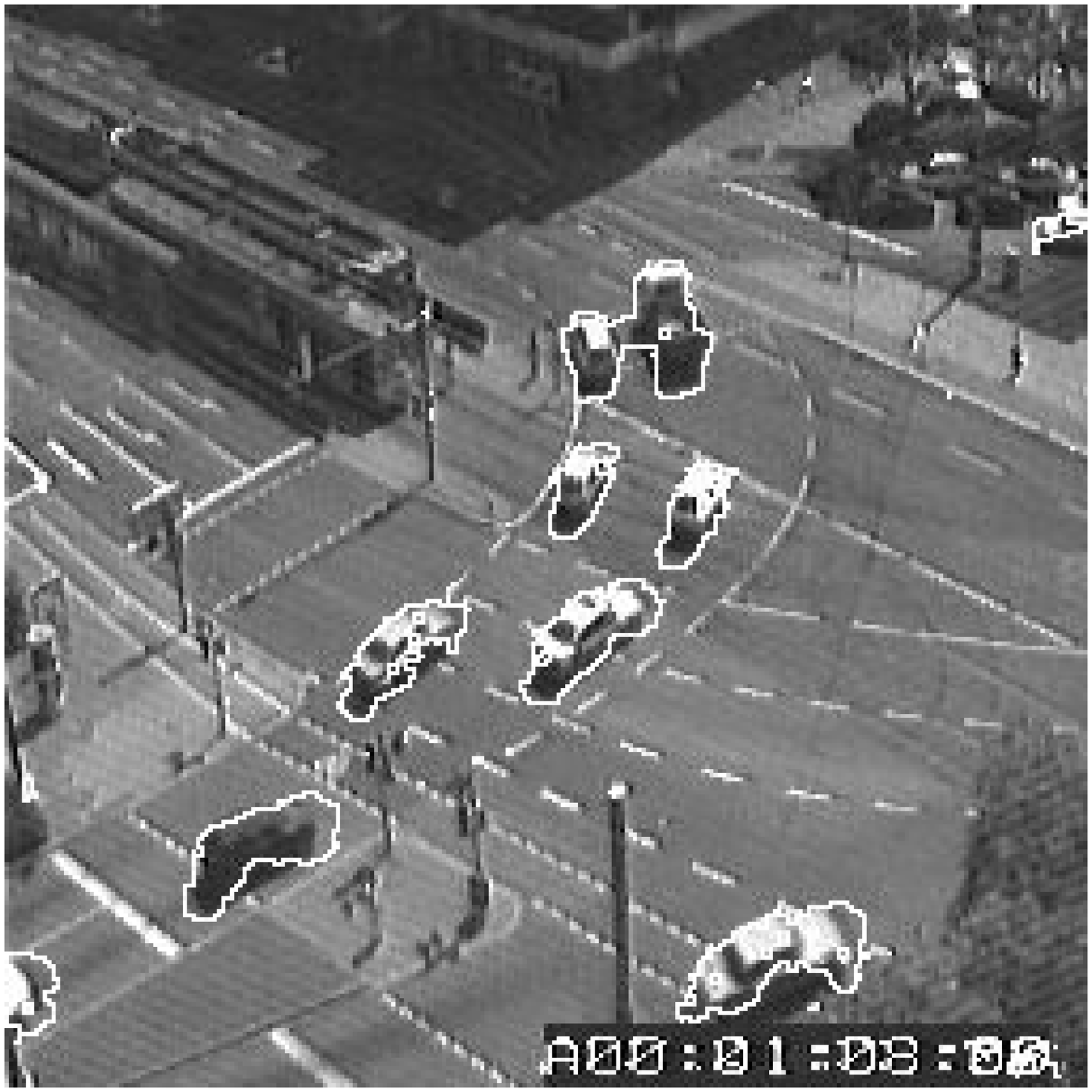}
\includegraphics[width=0.4\textwidth]{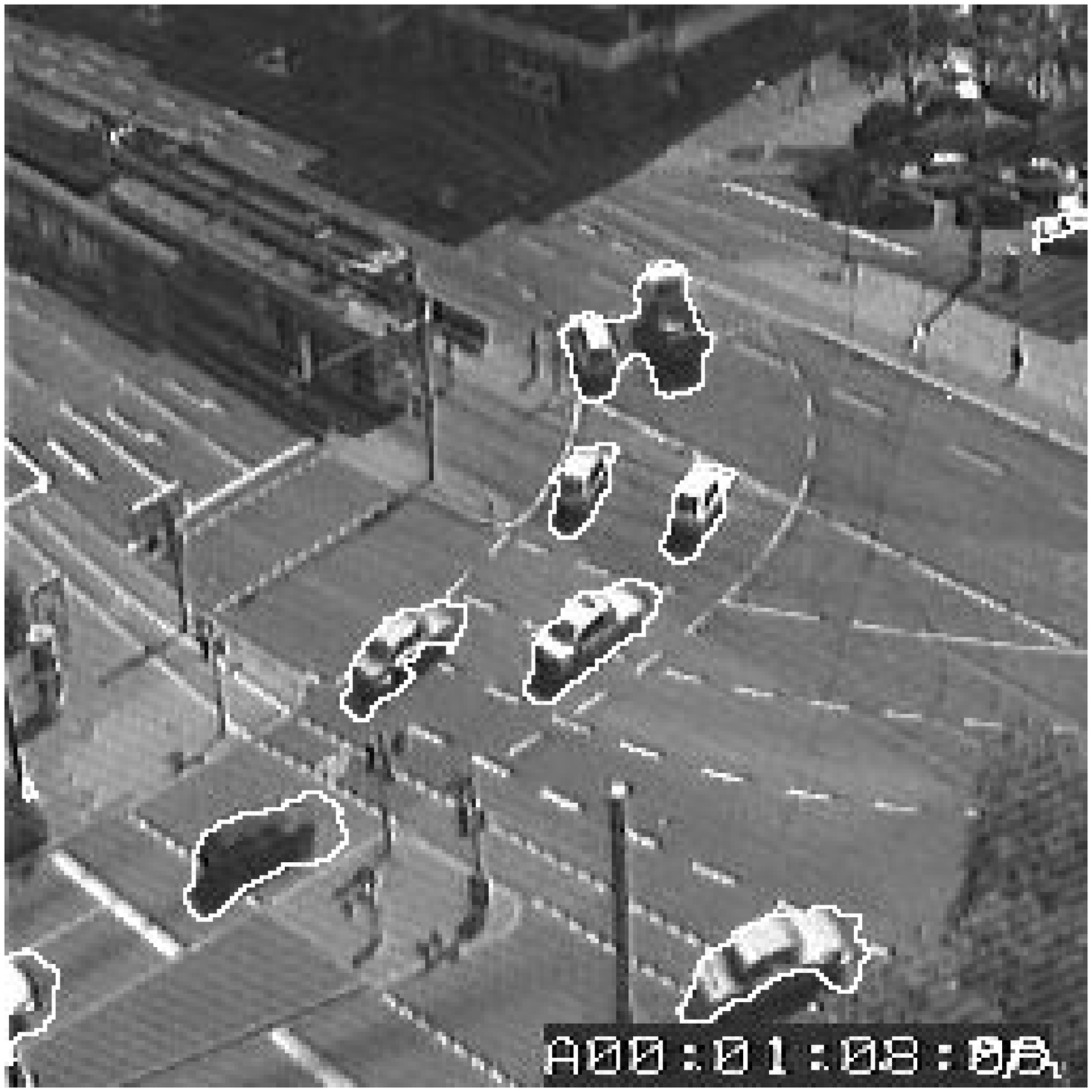}
\end{center}
\caption{{\small Results obtained (10th image of the sequence) with $TV_{1,g}$ and $TV_{1,g,\frac{\pi}{4}}$ and the optical flow norm as initial data. Parameters are $\alpha=0.6$,
    $\lambda=0.2$ and $\mu=10$. Notice the smoothness of the result on
    the right image in comparison to the one on the left image.}}
\label{fig:TV10}
\end{figure} 

\begin{figure}
\begin{center}
\includegraphics[width=0.4\textwidth]{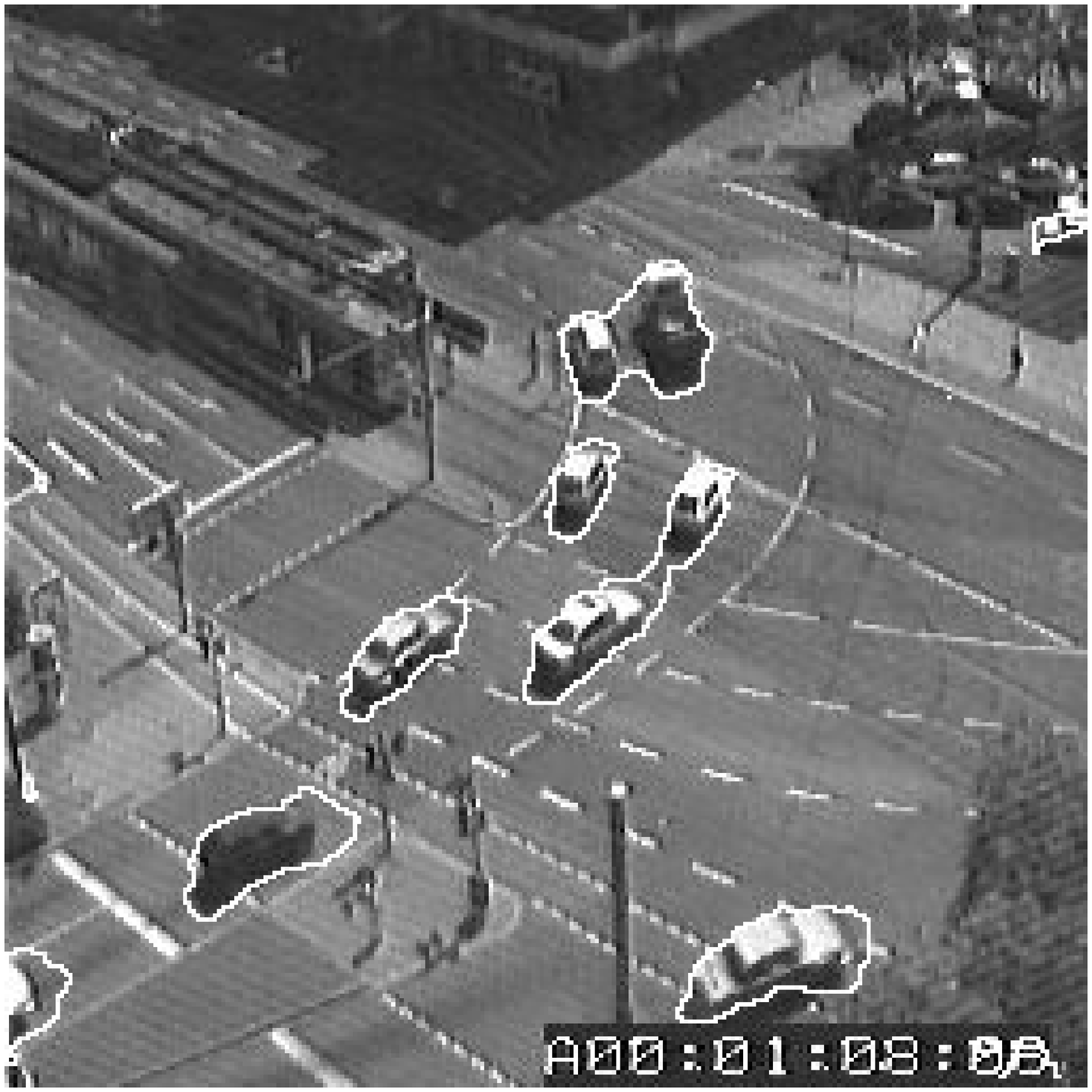}
\includegraphics[width=0.4\textwidth]{TV1mgseg_10.eps}
\includegraphics[width=0.4\textwidth]{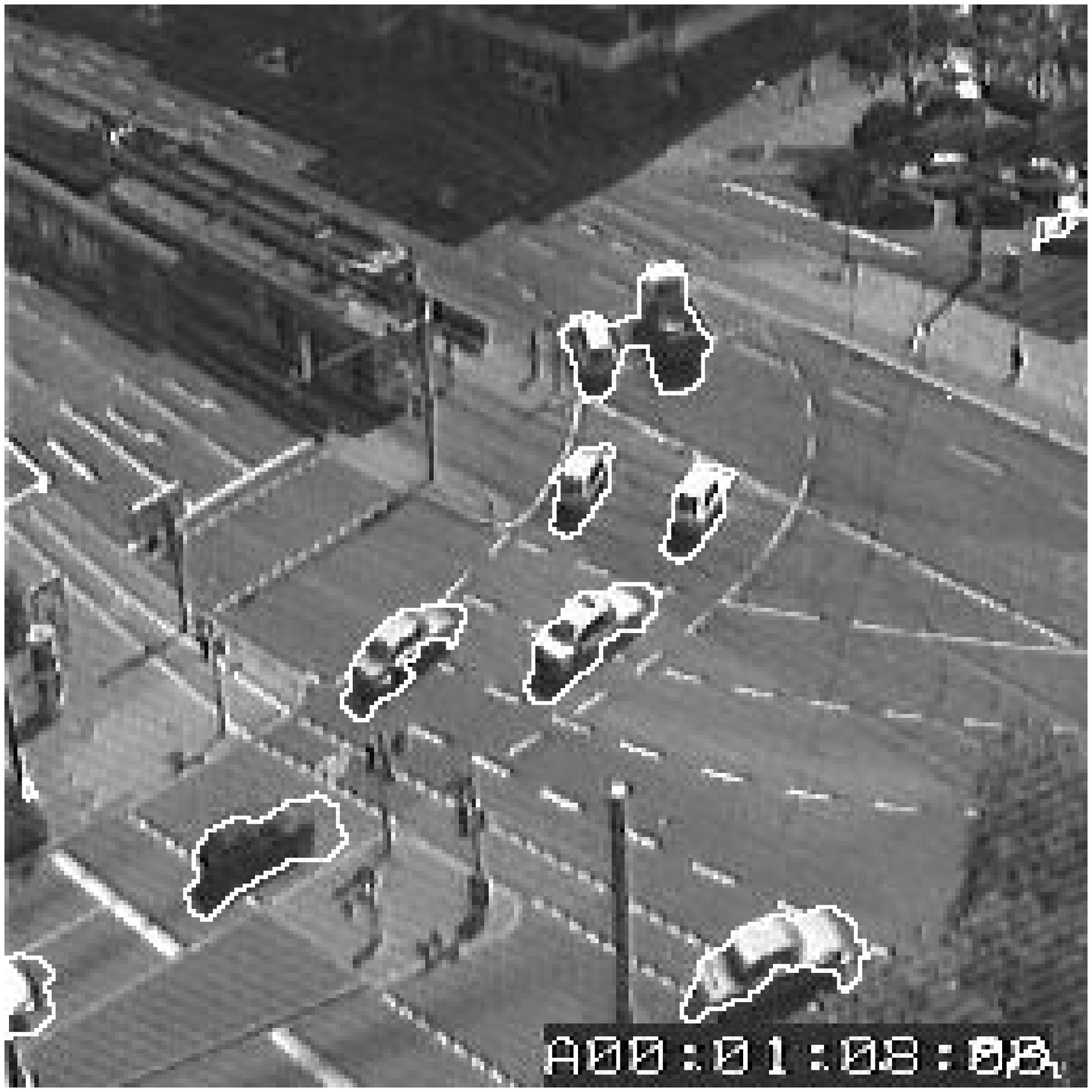}
\includegraphics[width=0.4\textwidth]{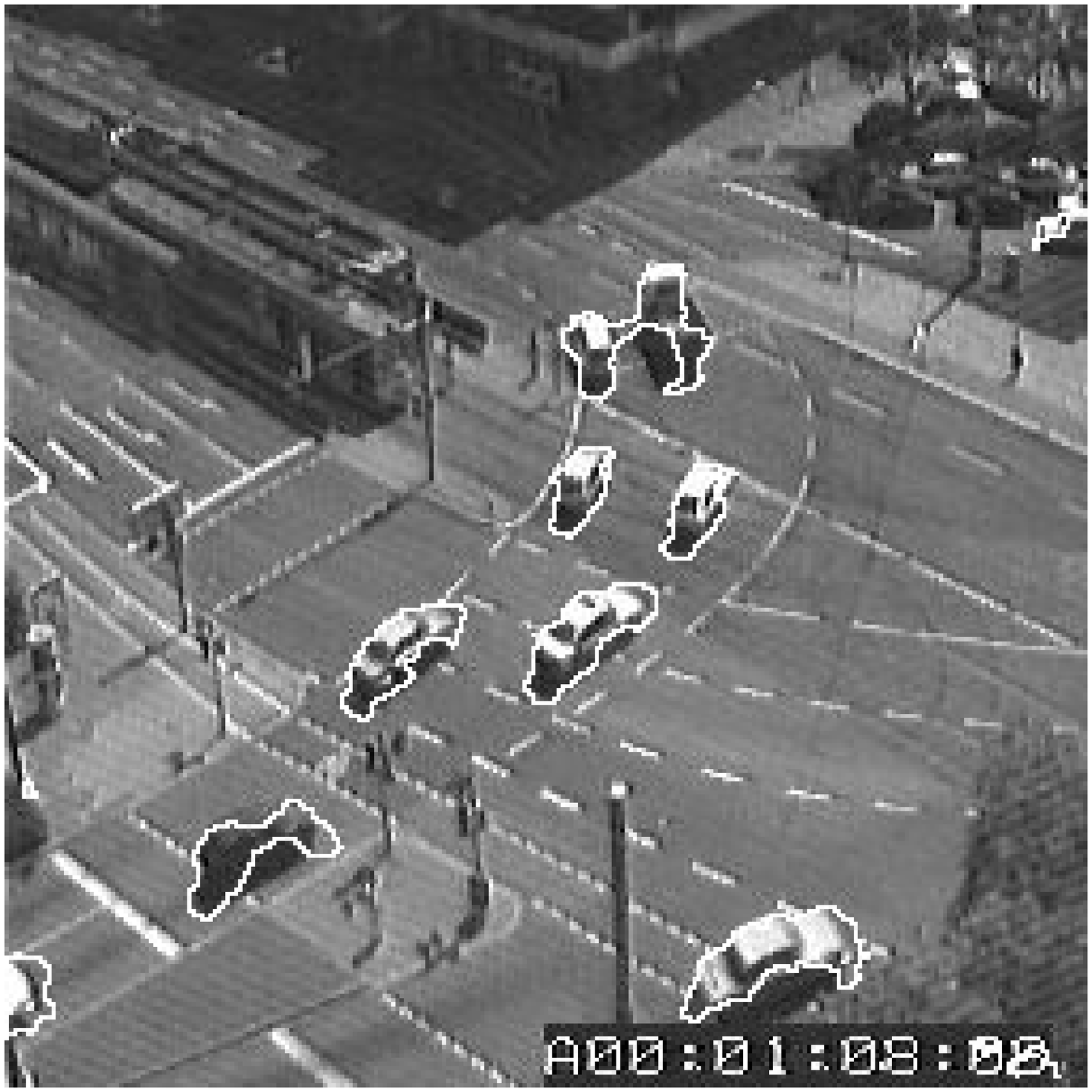}
\end{center}

\caption{{\small Influence of the $\alpha$ parameter. Results obtained
    (10th image of the sequence) with total variation minimisation
    with $TV_{1,g,\frac{\pi}{4}}$ and the optical flow norm as initial data. Parameters are $\lambda=0.2$ and $\mu=10$. From left to right and top to bottom~: $\alpha=0.5$, $0.6$, $0.7$ and $0.8$.}}
\label{fig:TVthresh}
\end{figure} 
The figure 2 shows the results obtained by solving the ROF problem and thresholding the function. We emphasize again that it is a major advantage over all previous way for solving this problem, since the function gives us all the solutions of the shape optimization problems depending on $\alpha$. The segmentation shown on figure 3 are indeed obtained simply by thresholding the function at the levels indicated ($0.5$, $0.7$ and $0.8$). As we had reasonable values of $\alpha$ from previous computations with classical snakes (\cite{ranchin}), we just tried a few values, but one could choose $\alpha$ in a more sophisticated way, adapted to the histogram of the function solving the ROF model. Such parameter optimization could also be applied in the same way to a functional that was used by Jehan-Besson, Barlaud and Aubert in \cite{AubBarJeh} for video segmentation purpose (actually it even inspired the work \cite{ranchin})
$$J(\Omega)=\int_{\Omega} \alpha\,dx+\int_{\Omega} |B-I|(x)\,dx+\lambda \int_{\partial\Omega}\,dS$$
where $B$ represent a background image and $I$ the current image in the movie. In the discrete formalism which is used in this paper, it gives
$$\sum_{i} (\alpha-|B-I|(i))\theta_{i}+\lambda TV_{1}(\theta).$$  
In this case it is a direct application of the previous work \cite{TV_MRF} (as before we have to modify the perimeter to a Manhattan perimeter). The background can be computed using more or less sophisticated methods. We tried time median filter and the method proposed by Kornprobst, Deriche and Aubert \cite{ADK}. Some results are shown on figure 4 for $\alpha=10, 15, 20, 25$ and $\lambda=50$.\\
\begin{figure}
\begin{center}
\includegraphics[width=0.4\textwidth]{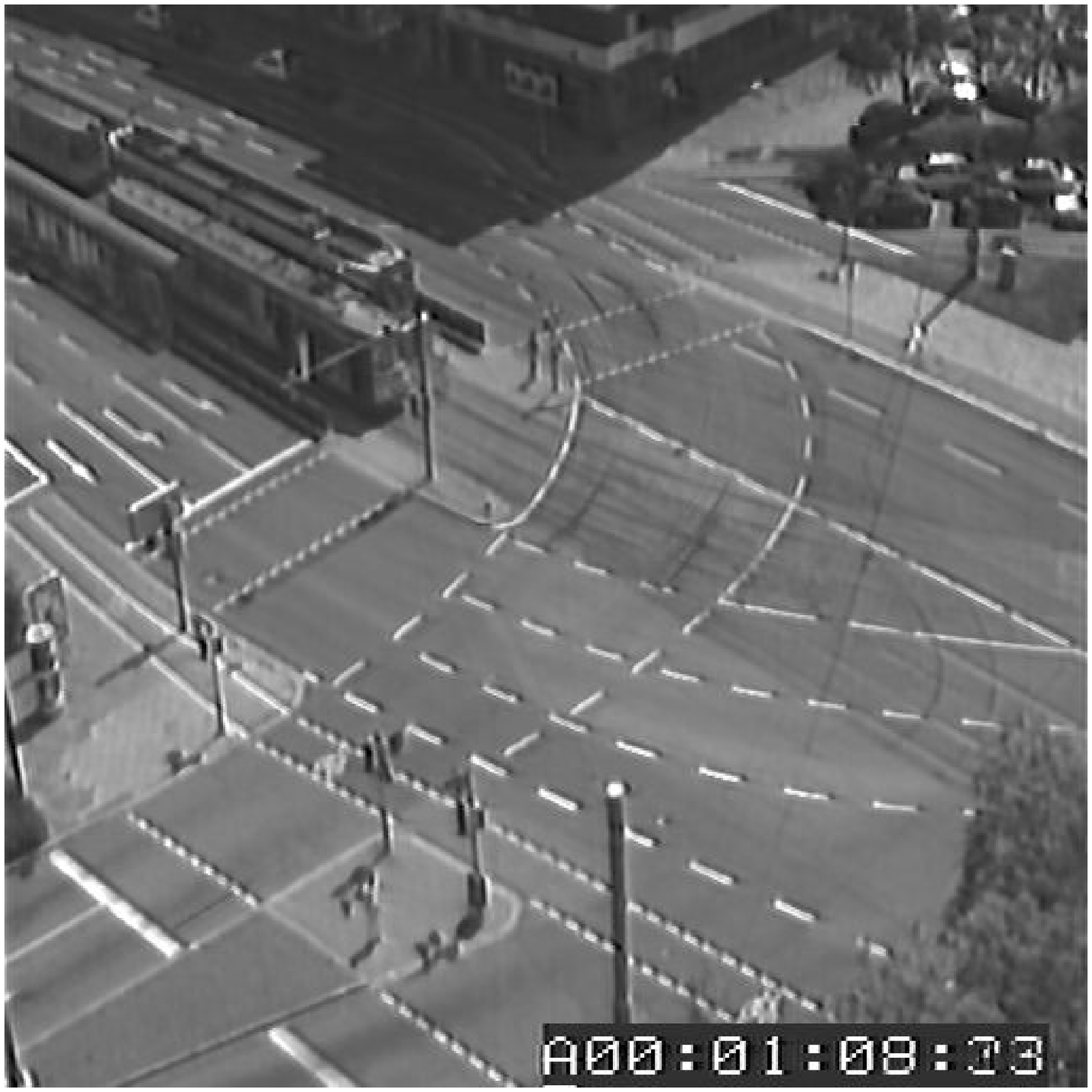}
\end{center}
\begin{center}
\includegraphics[width=0.4\textwidth]{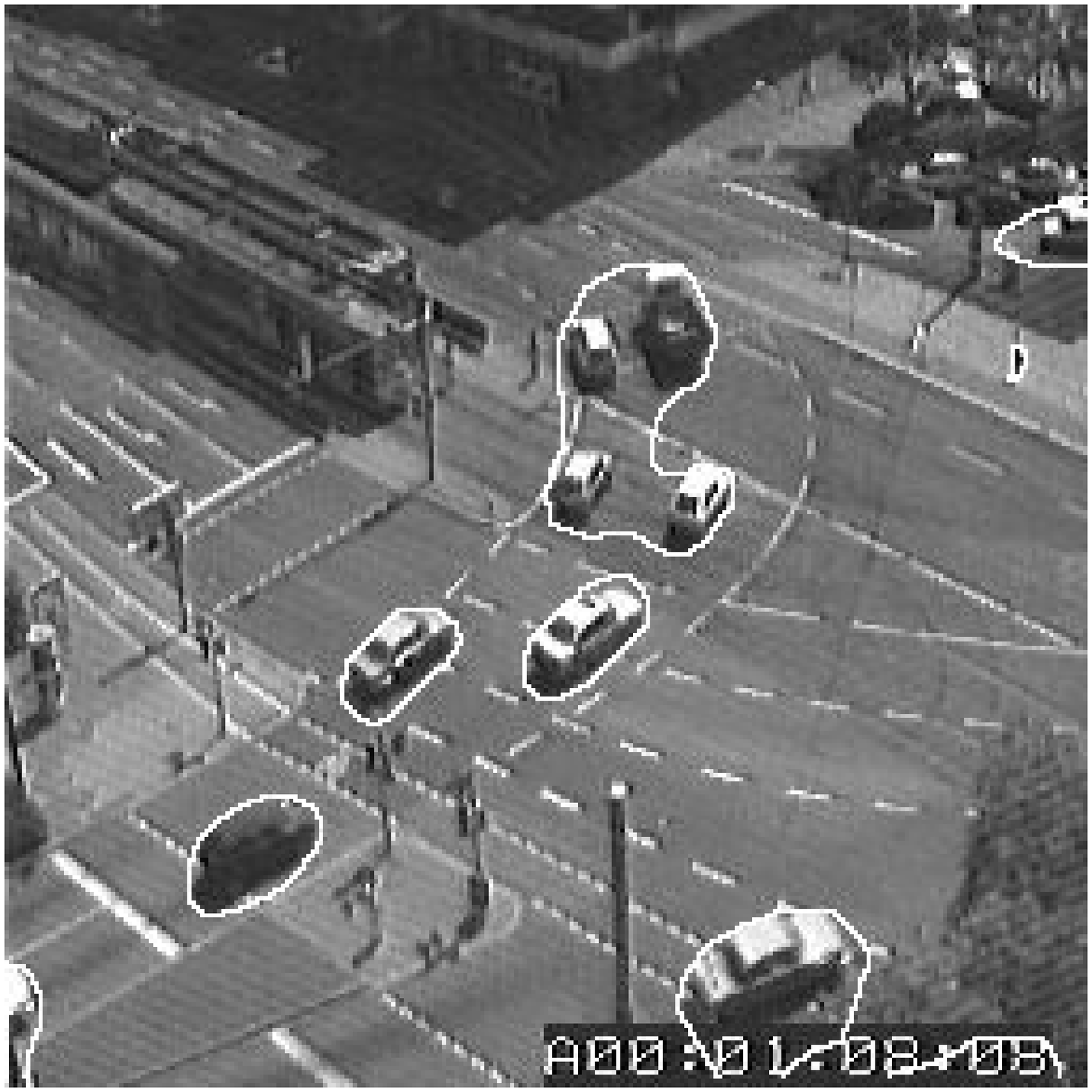}
\includegraphics[width=0.4\textwidth]{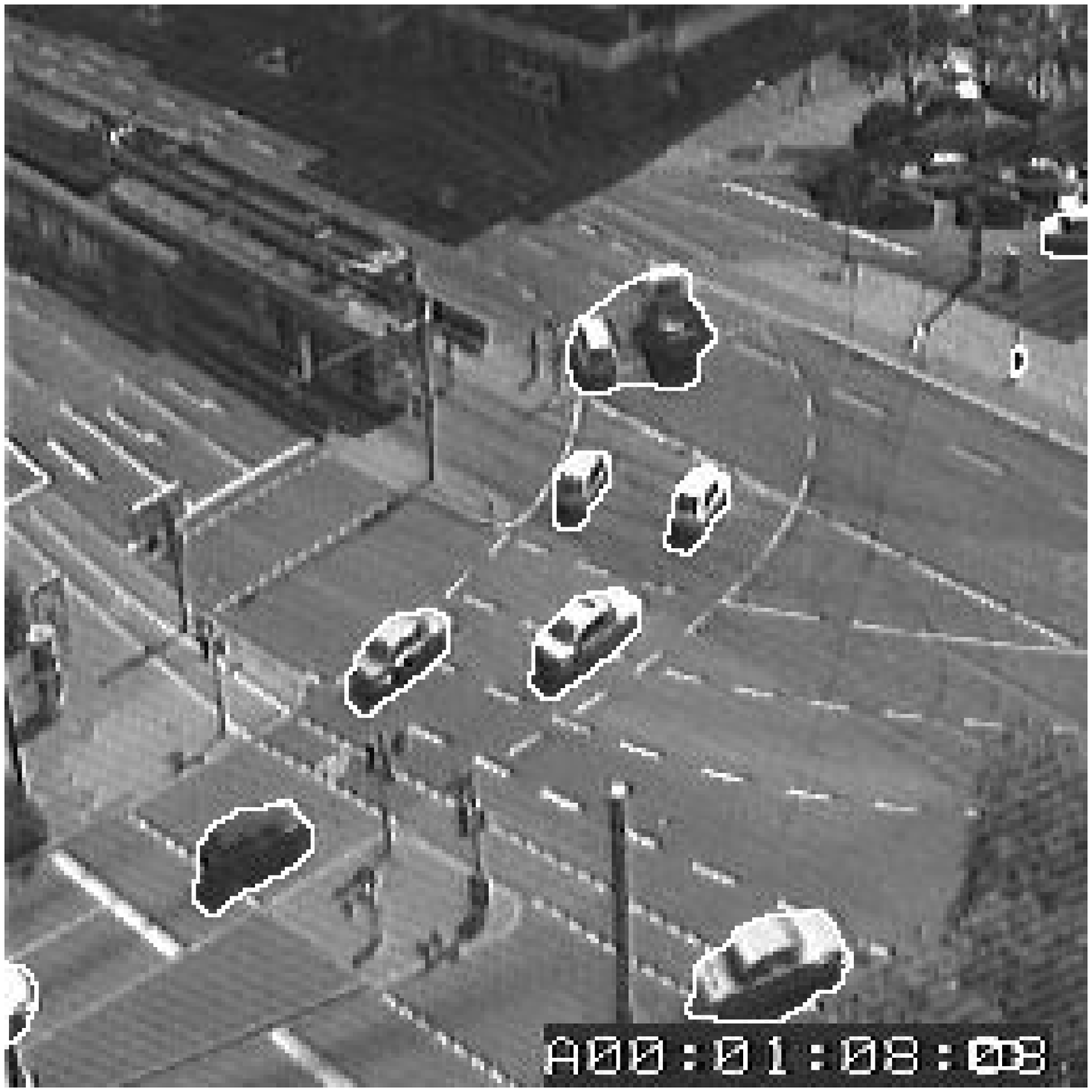}
\includegraphics[width=0.4\textwidth]{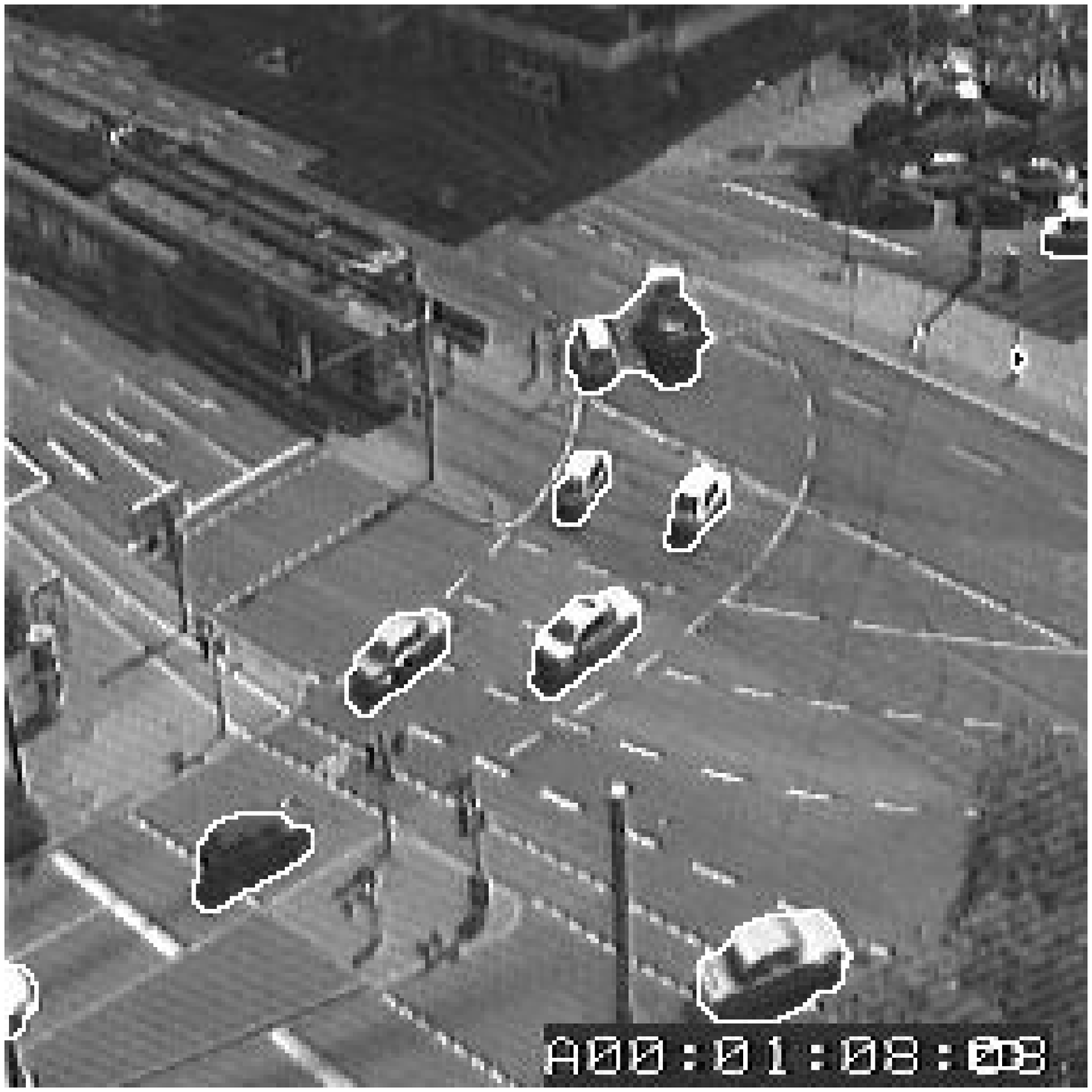}
\includegraphics[width=0.4\textwidth]{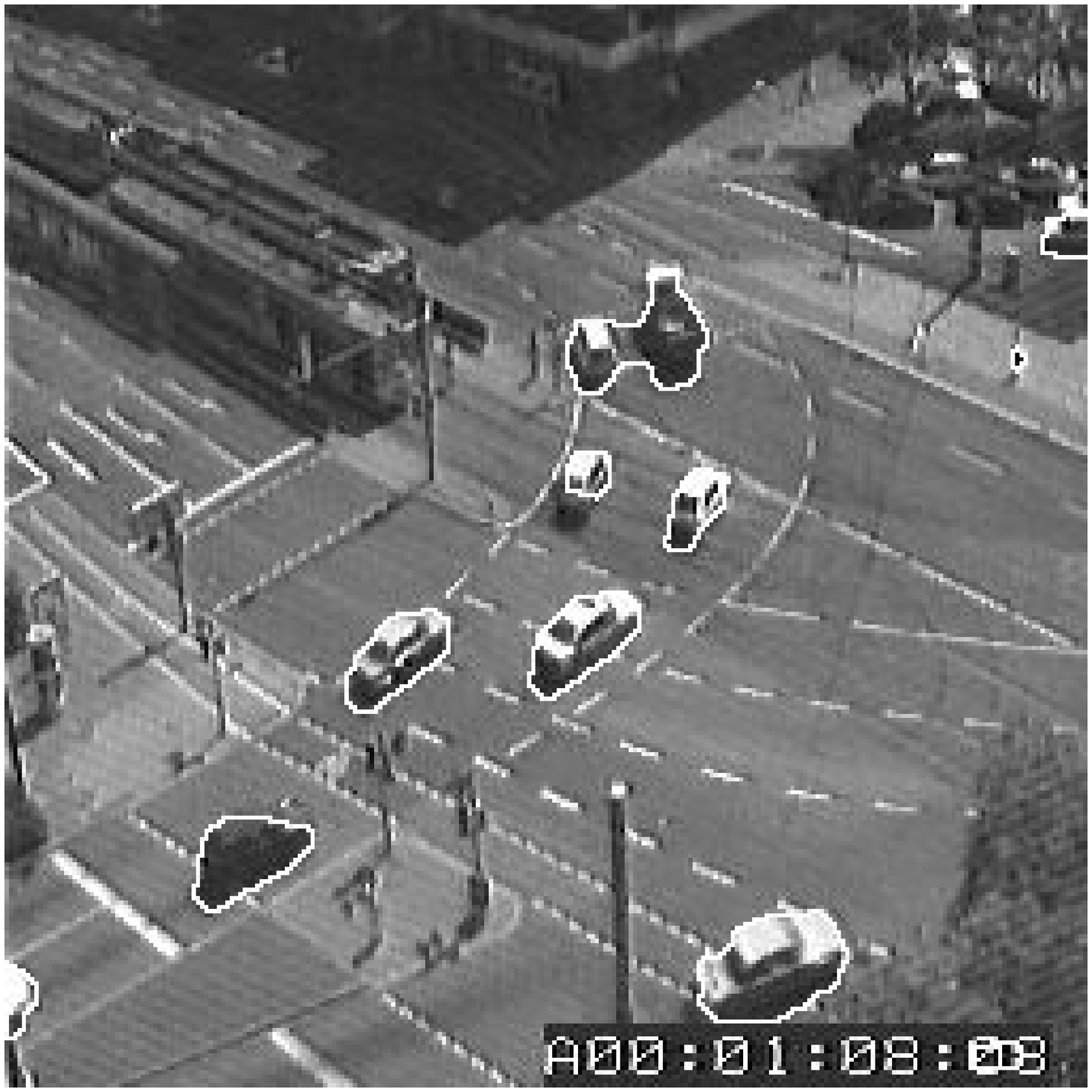}
\end{center}
\caption{{\small First image on the top~: background image computed by
    time median filter. Results obtained (10th image of the sequence)
    with total variation minimisation (Manhattan with horizontal,
    vertical and diagonal neighbors $TV_{1,g,\frac{\pi}{4}}$) for the
    Jehan-Besson--Aubert--Barlaud model (initial data: $|B-I|$). Parameters are $\lambda=50$. From left to right and top to bottom~: $\alpha=10, 15, 20, 25$.}}
\label{fig:imvsbg}
\end{figure}
Here is the computational time (measured in seconds with the \texttt{clock()} C command) of the algorithm (using the model described in \cite{ranchin}) on the ten first images of the sequence for the total variation minimisation algorithms (images are $256\times 256$, parameters are $\alpha=0.6$, $\lambda=0.2$ and $\mu=10$). Iterations were stopped when the maximum of the two residues between $\xi^{n}$ and $\xi^{n+1}$ and between $\eta^{n}$ and $\eta^{n+1}$ become lower than $0.002$, a maximum of $2000$ iterations being set to prevent the algorithm to become too slow. The time step is $\tau=0.1$. Such a value could be quite high, as we have indicated the time step should be lower than $\frac{1}{8\max_{i,j} g_{i,j}}$, but a simple trick is to write $g=\tilde{g}\max g$, which changes the regularization parameter from $1$ to $\max g$, and thus has no incidence over the time step condition, as this one does not depend on the regularization parameter. One could think the precision value is too low and leads to a quite heavy computational time, however, we have noticed that for a precision of $0.01$, the result is not sufficiently good for level sets extraction (see figure 5 where a result is displayed for precisions $0.01$ and $0.002$)
\begin{figure}
\begin{center}
\begin{tabular}{|c|c|c|}
\hline\\
 & \multicolumn{2}{c|}{Projection algorithm~:} \\
 & \multicolumn{2}{c|}{computational time with $TV_{1,g,\frac{\pi}{4}}$}\\\hline
time in seconds & iteration number & residue\\\hline
335.81 & 1739 & 0.001999\\
324.57 & 1722 & 0.001999\\
380.42 & 2001 & 0.002525\\
312.18 & 1691 & 0.002\\
314.08 & 1625 & 0.001999\\
330.00 & 1786 & 0.001999\\
379.77 & 2001 & 0.003226\\
312.70 & 1698 & 0.001999\\
371.66 & 2001 & 0.002088\\
\hline
\end{tabular}
\end{center}
\caption{{\small computational time of the TV regularization solving
    algorithm with $TV_{1,g,\frac{\pi}{4}}$. The residue $r=\max(\Vert \xi^{n+1}-\xi^{n}\Vert_{l^2},\Vert \eta^{n+1}-\eta^{n}\Vert_{l^2})$.}}
\end{figure}

\begin{figure}
\begin{center}
\begin{tabular}{|c|c|c|c|c|c|}
\hline\\
 & \multicolumn{2}{c|}{Projection algorithm~:} \\
 & \multicolumn{2}{c|}{Projection algorithm computational time with $TV_{1,g}$}\\\hline
time in seconds & iteration number & residue\\\hline
52.17 & 446 & 0.001998\\
61.32 & 530 & 0.001999\\
67.54 & 584 & 0.002000 \\
47.59 & 412 & 0.001999\\
49.50 & 429 & 0.001999\\
54.33 & 473 & 0.001999\\
66.56 & 553 & 0.001999\\
58.20 & 495 & 0.002000\\
56.25 & 461 & 0.001996\\
60.58 & 484 & 0.001999\\
\hline
\end{tabular}
\end{center}
\caption{{\small computational time of the TV regularization solving algorithm with $TV_{1,g}$. The residue $r=\max(\Vert \xi^{n+1}-\xi^{n}\Vert_{l^2},\Vert \eta^{n+1}-\eta^{n}\Vert_{l^2})$.}}
\end{figure}

\begin{figure}
\begin{center}
\includegraphics[width=0.4\textwidth]{TV1mgseg_10.eps}
\includegraphics[width=0.4\textwidth]{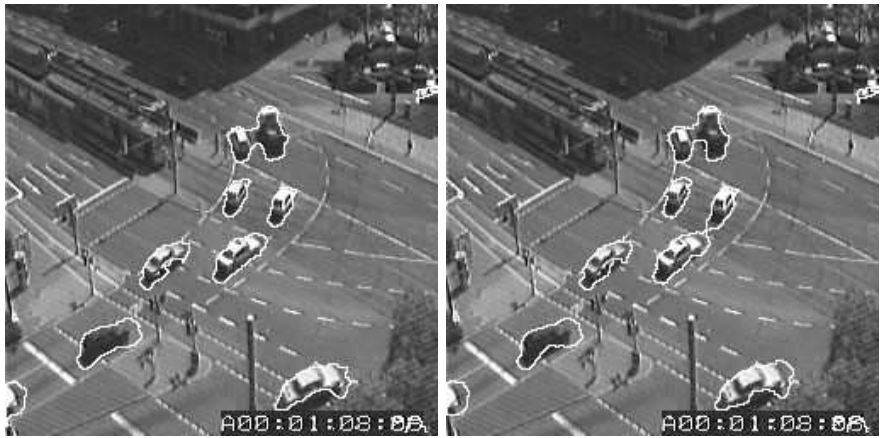}

\end{center}
\caption{{\small Results obtained (10th image of the sequence) with total variation minimisation (Manhattan with horizontal, vertical and diagonal neighbors) for two different precisions~: $0.002$ (left image) and $0.01$ (right image). Parameters are $\lambda=0.2$ and $\mu=10$.}}
\label{fig:precision}
\end{figure}


\section{Moving objects segmentation by \emph{a contrario} detection}
The method described here is inspired from previous works of Pelletier, Koepfler and Dibos \cite{sylvain} and Caselles, Garrido and Igual \cite{laura}. The purpose is to decide whether a pixel is \emph{meaningful} or not. In our case, the data is the solution of the ROF problem with the particular choice of the weighted total variation. The meaningfulness is decided between two hypothesis: $H^{0}$ \textquotedblleft there is motion\textquotedblright and $H^{1}$ \textquotedblleft there is no motion\textquotedblright.\\
The classical approach of hypothesis testing (\emph{hypothesis testing} model) is to suppose that $H^{0}$ is true and to have a look at the observations under this assumption. Another approach (\emph{a contrario} model) consists in deciding under the assumption that $H^{1}$ is true. This was introduced in \cite{DMM1} as a statistical method to provide a decision tool which simulates the Gestalt laws. The basic principle (\emph{Helmholtz} principle), is is based on the fact that every large deviation from the noise should be perceptible and thus is decided to be meaningful.\\
Around a pixel, we design a neighborhood $N(x)$ of size $N=n\times n$ and we define the random variable
$$\mathcal{E}_{x}=\frac{1}{N}\sum_{y\in N(x)} \psi(|\tilde{\mathcal{V}}(y)|)$$
where $\psi: \mathbb{R}\to [0,1]$  is a function designed to renormalize the data between zero and one. The pixels $\{y\in N(x)\}$ are assumed to be \textquotedblleft independent\textquotedblright and $\tilde{\mathcal{V}}$ denotes the random variable associated to the solution of the ROF problem with the optical flow amplitude as initial data.\\
Let $E_{x}$ the observed value of $\mathcal{E}_{x}$. There is motion if $E_{x}$ is sufficiently high. Then under the assumption that $H^{1}$ is true, the rejection test is $[\mathcal{E}_{x}\geq \delta],\ \delta>0$. But we do not compute the value of $\delta$ for a given level of meaningfulness as it is usually done in hypothesis testing, we compute the probability $\mathbb{P}[\mathcal{E}_{x}\geq E_{x}|H^{1}]$ which is the motion probability for the observed value $E_{x}$. For its evaluation, we need the Hoeffding's Theorem \cite{hoeffding}, once we have estimated the mean of the random variable $\psi(|\tilde{\mathcal{V}}(y)|)$ from the observed values.
\begin{thm} \textbf{(Hoeffding 1963)}
Let $Y^{1},...,Y^{N}$ be independent variables with $\mu^{i}=\mathbb{E}(Y^{i}) \in (0,1)$ and $\mathbb{P}[0\leq Y^{i}\leq 1]=1$ for all $i=1,...,N$. Let $\mu=\frac{\mu^{1}+...+\mu^{n}}{N}$. Then, for $0<t<1-\mu$ and $\bar{Y}=\frac{Y^{1}+...+Y^{N}}{N}$,
$$\mathbb{P}[\bar{Y}-\mu\geq t]\leq \exp(-NH(\mu+t,\mu))$$
where
$H(x,y)=x\log(\frac{x}{y})+(1-x)\log(\frac{1-x}{1-y})$
\end{thm}
Then we define the expected number of false alarms
\begin{defn} \textbf{(NFA of a pixel)}
The number of false alarms is defined as:
$$NFA(x)=\mathcal{N}_{tot}\mathbb{P}[\mathcal{E}_{x}\geq E_{x}|H^{1}]$$
where $\mathcal{N}_{tot}$ is the total number of pixels in the image.
\end{defn}
The rejection of $H^{1}$ is decided if the NFA is lower than a parameter $\epsilon$. For the estimation of $\mu$, we simply compute the empirical mean  of $E_{x}$ over the entire image:
$$\hat{\mu}=\frac{1}{\mathcal{N}_{tot}}\sum_{i} E_{x_{i}}.$$ 
Using the Hoeffding formula, a sufficient condition of rejection is then
$$H(E_{x},\hat{\mu})\geq\frac{1}{N}\log(\frac{\mathcal{N}_{tot}}{\epsilon})$$
for $\hat{\mu}<E_{x}<1$.\\
The main reservation of the application of this framework to the solution of the ROF problem is that the independency of this quantity over a neighborhood is not verified. However, we would like to emphasize that the dependency should exist only on the part of level lines included in the neighborhood. The TV regularization does not smooth accross the edges but along the edges. The second reason of this use of the Hoeffding formula is that practically, we do not notice any problem to apply this.\\
A post-treatment is done in order to take into account the fact that
the region detected should slightly surround the true motion region,
due to the neighborhood constructed around each pixel. We simply erode
the mask obtained by a radius of half the neighborhood radius. At the
end, we can compute the level set of the ROF problem solution which
has the minimal difference with the result obtained with the \emph{a
  contrario} detection.\\
We present results on figures 6 and 7. On the
figure 6 (\emph{resp.} 7), the observation is the result of the ROF problem with
weighted TV and optical flow norm as initial data (\emph{resp.}
difference image $B-I$); on the left image
is the result obtained from the \emph{a contrario} detection (plus
erosion), the closer level set is shown on the right image. We can
notice the \emph{a contrario} method is not able to discriminate
between two moving cars in the image.
\begin{figure}
\begin{center}
\includegraphics[width=0.4\textwidth]{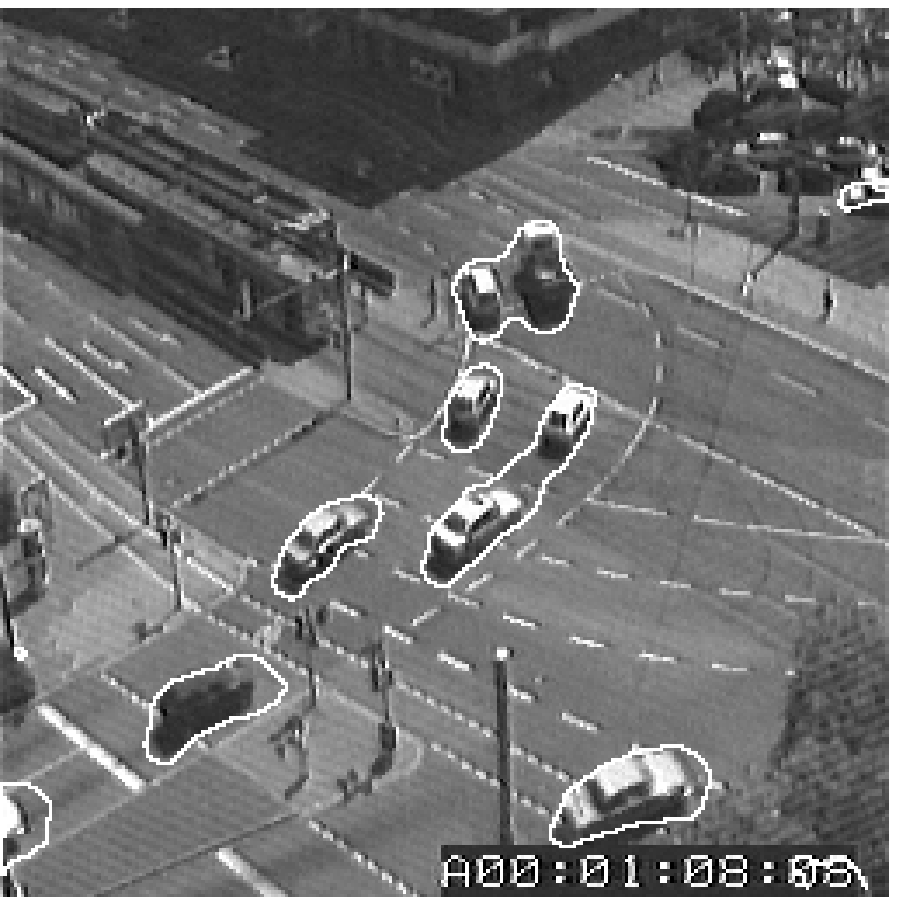}
\includegraphics[width=0.4\textwidth]{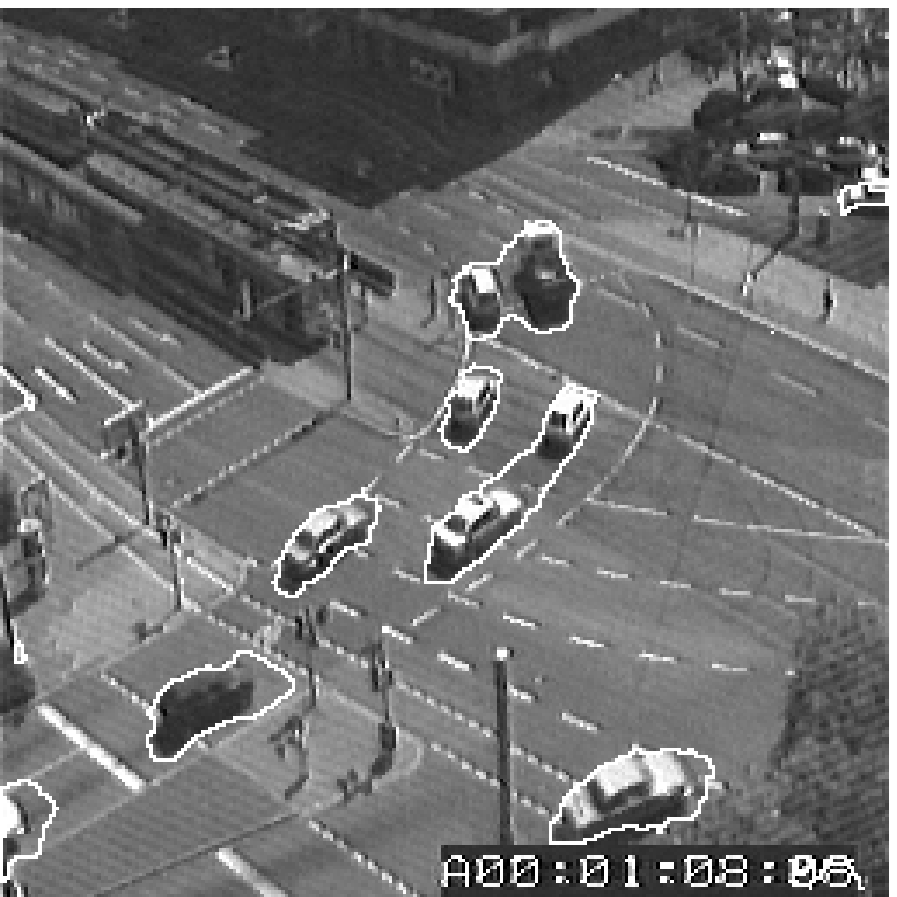}
\caption{{\small \emph{a contrario} detection with the optical flow magnitude regularized by $TV_{1,g,\frac{\pi}{4}}$. The neighborhood radius is $N=3$. The $\epsilon$ parameter is set to one as it is usually done. The left image is the basic result of the \emph{a contrario} detection eroded with a radius of $1$. The right image is one level set of the ROF solution which looks like best the \emph{a contrario} detection result.}}
\end{center}
\label{fig:nfa}
\end{figure}

\begin{figure}
\begin{center}
\includegraphics[width=0.4\textwidth]{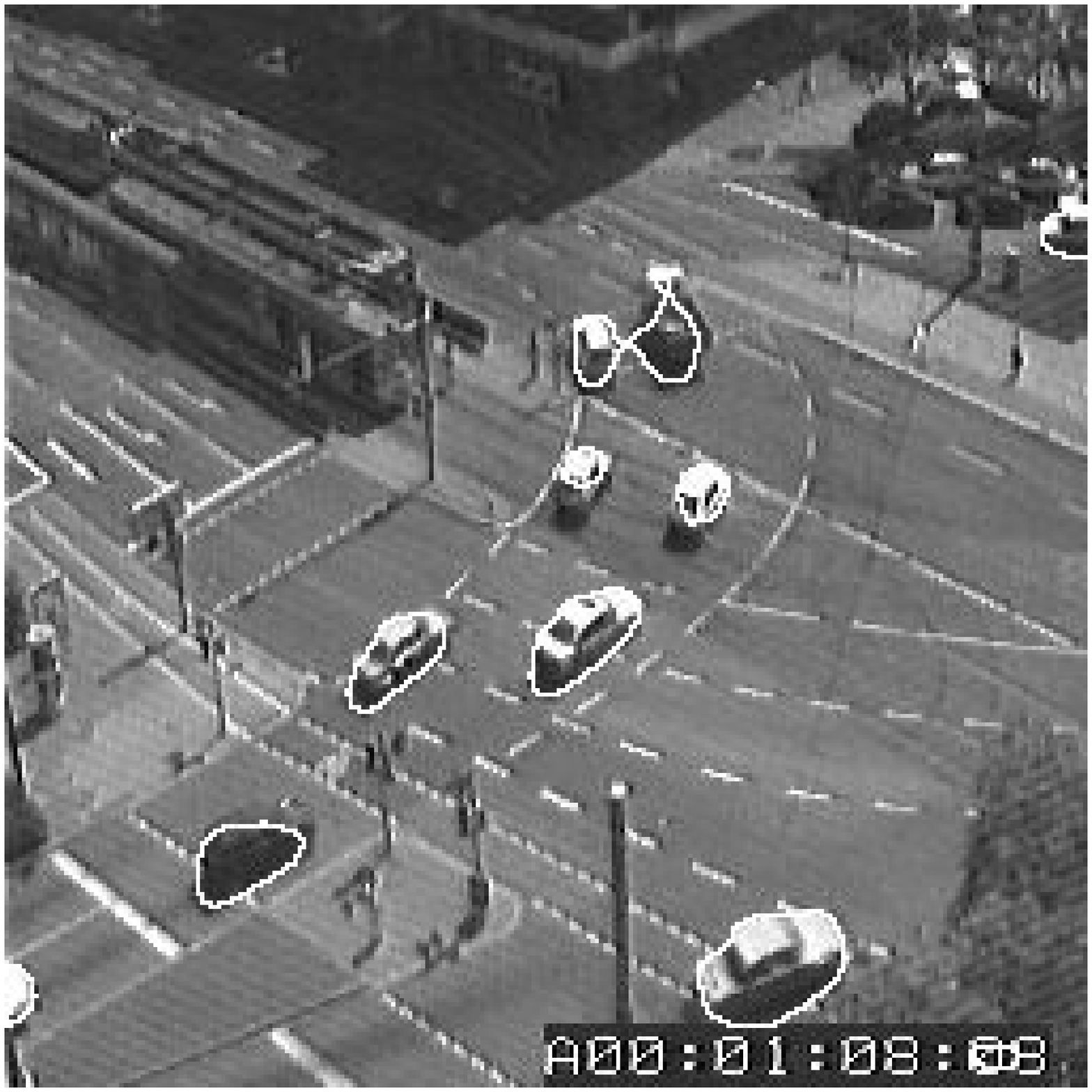}
\includegraphics[width=0.4\textwidth]{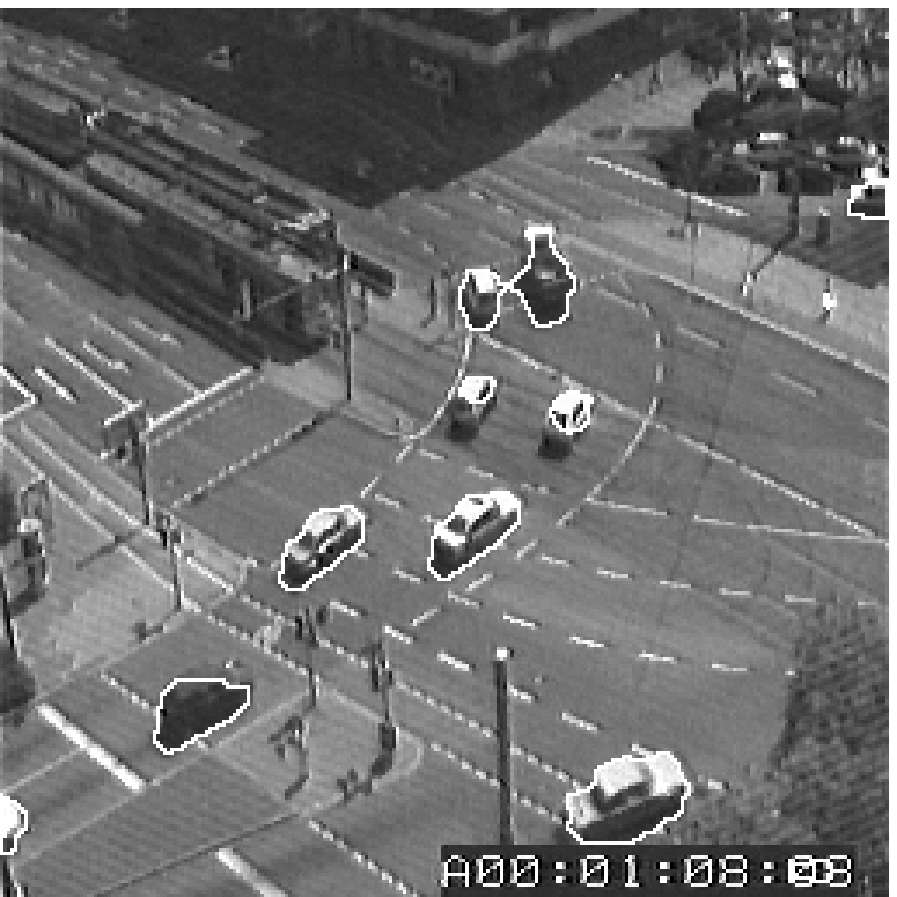}

\caption{{\small \emph{a contrario} detection with the difference image between the current image and the background regularized by $TV_{1,g,\frac{\pi}{4}}$. The neighborhood radius is $N=3$. The $\epsilon$ parameter is set to one as it is usually done. The left image is the basic result of the \emph{a contrario} detection eroded with a radius of $1$. The right image is one level set of the ROF solution which looks like best the \emph{a contrario} detection result.}}
\end{center}
\label{fig:nfa2}
\end{figure}

\section{Conclusion}
In this paper, we have extended the main result of \cite{TV_MRF} in order to handle shape optimization functionals involving weighted anisotropic perimeter. It states that all the solutions of some shape optimization problems depending on a parameter $\alpha$ are $\alpha$-level sets of the solution of the Rudin-Osher-Fatemi problem. Thus the algorithm used for total variation regularization~---~as in \cite{TV_MRF}~---~allows to compute all the solutions for different values of $\alpha$ in one pass. This is in our opinion the main advantage over classical snakes methods like Chan and Vese one in this particular type of shape optimization.\\
On the other hand, we have also minimized the discrete version of the functional with graph cuts techniques. The main advantage of this method is that it is very fast, and whenever the advantage of the TV-minimization algorithm does not occur when we employ graph cuts, even a great number of computations of the algorithm leads to a very competitive computational time (close to a single computation of a classical continuous snake algorithm).\\
We have used these both methods on two video segmentation models~: one introduced in \cite{ranchin} in which weighted perimeter is involved and a previous one introduced by Jehan-Besson, Barlaud and Aubert \cite{AubBarJeh}. We would like to emphasize that the general model studied in the theoretical part of the paper covers many applications. One could think for example about segmentation with shape priors, using a perimeter weighted by a distance to the prior. Such models have been used by Freedman and Zhang \cite{Freedman}, or by Gastaud, Jehan-Besson, Barlaud and Aubert \cite{ABJBG}...\\
We have also proposed o use an \emph{a contrario} method for region finding with the result of TV minimization process but without extracting a level set at a predefinite level. This method do not lead to the most satisfactory results, but it is very fast since it does not require to choose a value of the parameter in particular and thus is better indicated for real-time applications.
\bibliography{biblio}

\begin{thebibliography}{10}

\bibitem{DMM1}
L.~Moisan A.~Desolneux and J.-M. Morel.
\newblock Edge detection by hemholtz principle.
\newblock {\em Journal of Mathematical Imaging and Vision}, 14:271--284, 2001.

\bibitem{AubBarJeh}
G.~Aubert, M.~Barlaud, and S.~Jehan-Besson.
\newblock Video object segmentation using eulerian region-based active
  contours.
\newblock In {\em International Conference in Computer Vision Proceedings},
  Vancouver, Canada, 2001.

\bibitem{ADK}
G.~Aubert, R.~Deriche, and P.~Kornprobst.
\newblock Image sequence analysis via partial differential equations.
\newblock {\em Journal of Mathematical Imaging and Vision}, 11(1):5--26, 1999.

\bibitem{morpho}
L.~Bouchard, I.~Corset, S.~Jeannin, F.~Marqués, F.~Meyer, R.~Morros, M.~Pardàs,
  B.~Marcotegui, and P.~Salembier.
\newblock Segmentation-based video coding system allowing the manipulation of
  objects.
\newblock {\em IEEE Transactions on Circuits and Systems for Video Technology,
  (RACE/MAVT and MORPHECO Projects)}, 7(1):60--74, 1997.

\bibitem{PHB}
P.~Bouthemy, F.~Heitz, and P.~Pérez.
\newblock Multiscale minimization of global energy functions in some visual
  recovery problems.
\newblock {\em CVGIP~: Image understanding}, 59(1):191--212, 1989.

\bibitem{BL}
P.~Bouthemy and P.~Lalande.
\newblock Recovery of moving object masks in an image sequence using local
  spatiotemporal contextual information.
\newblock {\em Optical Engineering}, 32(6):1205--1212, 1993.

\bibitem{gc_med}
Y.~Boykov and M.-P. Jolly.
\newblock Interactive organ segmentation using graph cuts.
\newblock {\em Medical Image Computing and Computer-Assisted Intervention},
  pages 276--286, 2000.

\bibitem{graphcuts}
Y.~Boykov, O.~Veksler, and R.~Zabih.
\newblock Fast approximate energy minimization via graph cuts.
\newblock {\em IEEE Transactions on Pattern Analysis and Machine Intelligence},
  23(11):1222--1239, 2001.

\bibitem{CKS}
V.~Caselles, R.~Kimmel, and G.~Sapiro.
\newblock Geodesic active contours.
\newblock {\em International Journal of Computer Vision}, pages 694--699, 1995.

\bibitem{tv_min}
A.~Chambolle.
\newblock An algorithm for total variation minimization and applications.
\newblock {\em Journal of Mathematical Imaging and Vision}, 20(1-2):89--97,
  2004.

\bibitem{TV_MRF}
A.~Chambolle.
\newblock Total variation minimization and a class of binary mrf models.
\newblock In Anand Rangarajan, Baba Vemuri, and Alan~L. Yuille, editors, {\em
  Proceedings of the 5th International Workshop on Energy Minimization Methods
  in Computer Vision and Pattern Recognition}, volume 3757 of {\em LNCS}, pages
  136--152, 2005.

\bibitem{CV}
T.~F. Chan and L.~A. Vese.
\newblock Active contours without edges.
\newblock {\em IEEE Transactions on Image Processing}, 10(2):266--277, 2001.

\bibitem{Nikolova}
T.F. Chan, S.~Esedoglu, and M.~Nikolova.
\newblock Algorithms for finding global minimizers of image segmentation and
  denoising models.
\newblock Technical Report CAM 04-07, UCLA, February 2004.

\bibitem{DS1}
J.~Darbon and M.~Sigelle.
\newblock Exact optimization of discrete constrained total variation
  minimization problems.
\newblock In R.~Klette and J.~Zunic, editors, {\em Tenth International Workshop
  on Combinatorial Image Analysis}, volume 3322 of {\em LNCS}, pages 548--557,
  December 2004.

\bibitem{DS2}
J.~Darbon and M.~Sigelle.
\newblock A fast and exact algorithm for total variation minimization.
\newblock In J.~S. Marques, N.~P\'erez de~la Blanca, and P.~Pina, editors, {\em
  2nd Iberian Conference on Pattern Recognition and Image Analysis}, volume
  3522 of {\em LNCS}, pages 351--359, June 2005.

\bibitem{DelZol}
M.~Delfour and J.-P. Zolésio.
\newblock {\em Shapes and Geometries}.
\newblock SIAM, Philadelphia, PA, Advances in Design and Control, 2001.

\bibitem{DerPar}
R.~Deriche and N.~Paragios.
\newblock Geodesic active regions for motion estimation and tracking.
\newblock {\em Proceedings of the Int. Conf. in Computer Vision}, pages
  224--240, 1999.

\bibitem{francoise}
F.~Dibos and G.~Koepfler.
\newblock Global total variation minimization.
\newblock {\em SIAM Journal of Numerical Analysis}, 37:646--664, 2000.

\bibitem{francoise2}
F.~Dibos, G.~Koepfler, and P.~Monasse.
\newblock {\em Total Variation Minimization: Application to Gray-Scale, Color
  Images and Optical Flow Regularization}.
\newblock Springer, 2003.

\bibitem{GPS}
B.T.~Porteous D.M.~Greig and A.H. Seheult.
\newblock Exact maximum a posteriori estimation for binary images.
\newblock {\em J. R. Statist. Soc. B}, 25:271--279, 1989.

\bibitem{sylvain}
G.~Koepfler F.~Dibos and S.~Pelletier.
\newblock Real-time segmentation of moving objects in a video sequence by a
  contrario detection.
\newblock In {\em Proceedings of the International Conference of Image
  Processing}, 2005.

\bibitem{Freedman}
D.~Freedman and T.~Zhang.
\newblock Interactive graph cut based segmentation with shape priors.
\newblock {\em IEEE Computer Society Conference on Computer Vision and Pattern
  Recognition (CVPR)}, 1:755--762, 2005.

\bibitem{ABJBG}
M.~Gastaud, S.~Jehan-Besson, M.~Barlaud, and G.~Aubert.
\newblock Region-based active contours using geometrical and statistical
  features for image segmentation.
\newblock In {\em Proceedings of the IEEE International Conference in Image
  Processing}, volume~II, pages 643--646, 2003.

\bibitem{hoeffding}
W.~Hoeffding.
\newblock Probability inequalities for sums of bounded random variables.
\newblock {\em Journal of the American Statistical Association}, 58:13--30,
  1963.

\bibitem{gc_klm}
V.~Kolmogorov and R.~Zabih.
\newblock What energy functions can be minimized via graph cuts?
\newblock {\em IEEE Transactions on Pattern Analysis and Machine Intelligence
  (PAMI)}, 2004.

\bibitem{gc_stereo}
V.~Kolmogorov, R.~Zabih, and S.~Gortler.
\newblock Generalized multi-camera scene reconstruction using graph cuts.
\newblock {\em EMMCVPR 03 Proceedings}, 2003.

\bibitem{MP}
E.~M\'emin and P.~Pérez.
\newblock A multigrid approach for hierarchical motion estimation.
\newblock In {\em Proceedings of the 6th International Journal of Computer
  Vision}, pages 933--938. IEEE Computer Society Press, 1998.

\bibitem{MS}
D.~Mumford and J.~Shah.
\newblock Optimal approximations by piecewise smooth functions and associated
  variational problems.
\newblock {\em Communications on Pure and Applied Mathematics}, 42:577--684,
  1989.

\bibitem{MuratSimon}
F.~Murat and J.~Simon.
\newblock Sur le contr\^ole par un domaine g\'eom\'etrique.
\newblock Pr\'e-publication du Laboratoire d'Analyse Num\'erique, no 74015,
  Universit\'e de Paris 6, 222 pages.

\bibitem{OS}
S.~Osher and J.~Sethian.
\newblock Fronts propagating with curvature dependent speed: Algorithms based
  on the hamilton-jacobi formulation.
\newblock {\em Journal of Computational Physics}, 79:12--49, 1990.

\bibitem{ranchin}
F.~Ranchin and F.~Dibos.
\newblock Segmentation des objets en mouvement par utilisation du flot optique.
\newblock {\em ORASIS 2005 Proceedings}, 2005.

\bibitem{ROF}
L.~Rudin, S.~Osher, and E.~Fatemi.
\newblock Non linear total variation based noise removal algorithms.
\newblock {\em Physica D}, 60:259--268, 2002.

\bibitem{zolesio}
J.~Sokolowski and J.-P. Zolésio.
\newblock {\em Introduction to Shape Optimization. Shape sensitivity analysis}.
\newblock Springer Ser. Comput. Math. Springer-Verlag, 1992.

\bibitem{laura}
L.~Garrido V.~Caselles and L.~Igual.
\newblock A contrast invariant approach to motion estimation.
\newblock In {\em Proceedings of the International Conference on Scale Space
  2005}, 2005.

\bibitem{WS}
J.~Weickert and C.~Schnörr.
\newblock Variational optic flow computation with a spatio-temporal smoothness
  constraint.
\newblock {\em Journal of Mathematical Imaging and Vision}, 14:245--255, 2001.

\end{thebibliography}
\bibliographystyle{plain}

\end{document}